\pdfoutput=1

\documentclass[11pt]{article}

\usepackage[final]{acl}

\usepackage{times}
\usepackage{latexsym}

\usepackage[T1]{fontenc}

\usepackage[utf8]{inputenc} 
\usepackage[T1]{fontenc}    

\PassOptionsToPackage{hyperindex,breaklinks,unicode}{hyperref}

\usepackage{url}
\usepackage{booktabs}
\usepackage{hyperref}
\usepackage{graphicx}
\usepackage{microtype}
\usepackage{multicol}
\usepackage{xcolor}   
\usepackage{multirow}
\usepackage{lipsum}
\usepackage{enumitem}
\usepackage{pifont}
\usepackage{colortbl}
\usepackage{tabularx}
\usepackage{caption} 
\usepackage{float}
\usepackage{amsmath,amsfonts,amssymb,bbm}

\usepackage{lineno}
\usepackage{makecell}
\usepackage{array}
\usepackage{wrapfig}
\usepackage{tikz}
\usepackage[normalem]{ulem}
\usepackage{marvosym}
\usepackage{tcolorbox}
\tcbuselibrary{skins, breakable, theorems}
\usepackage{subcaption}  
\usepackage{float}  
\usepackage{listings}
\usepackage{fontawesome5}

\usepackage{CJKutf8}

\definecolor{gold}{RGB}{212,175,55}
\definecolor{silver}{RGB}{192,192,192}
\definecolor{bronze}{RGB}{205,127,50}
\definecolor{worst}{RGB}{200,40,40}

\newcommand{\gold}[1]{\textbf{\textcolor{gold}{\faTrophy\ #1}}}
\newcommand{\silver}[1]{\textbf{\textcolor{silver}{\faMedal\ #1}}}
\newcommand{\bronze}[1]{\textbf{\textcolor{bronze}{\faAward\ #1}}}

\definecolor{deepred}{rgb}{0.6, 0, 0}

\title{On the Systematic Challenges of Culturally Loaded Machine Translation:\\ \textbf{\textit{Dream of the Red Chamber}} as the Cultural Lens}

\author{
  Yiming Wang$^*$~~~~~~~~~~~~~ \\
  School of Computer Science,~~~~~~~~~~~~~ \\Shanghai Jiao Tong University~~~~~~~~~~~~~\\
  \texttt{yiming.wang@sjtu.edu.cn}~~~~~~~~~~~~~
  \And Jiayuan Di\thanks{Equal Contribution. We also thank Natsuki Oe, Kanon Yamaguchi, and Mouye Weng for coordinating the human study with Japanese volunteers, and Yuya Goto of Mejiro University for his suggestions on parts of the background discussion and case studies in this manuscript.} \\
  School of Foreign Languages, \\East China University of Science and Technology \\
  \texttt{mira.jiayuan@gmail.com}\\
}

\begin{document}
\maketitle

\begin{abstract}
Culturally loaded translation poses unique challenges for machine translation (MT), as meanings are deeply embedded in socio-cultural contexts beyond surface linguistic forms.
Although large language models (LLMs) have enabled MT systems to achieve human-like quality in many scenarios, their ability to handle culturally loaded expressions remains underexplored.
In this study, we systematically investigate the challenges posed by culturally loaded translation in LLM-based MT systems.
We construct a Chinese-Japanese bilingual dataset from the culturally representative corpus {\it Dream of the Red Chamber}\footnote{The Chinese novel \begin{CJK}{UTF8}{gbsn}《红楼梦》\end{CJK}, also known as \begin{CJK}{UTF8}{gbsn}《石头记》\end{CJK}, was first translated as \textit{Dream of the Red Chamber} by Wang Jizhen (1929). David Hawkes later adopted the alternative title \textit{The Story of the Stone} in his influential translation (1973–1986). Following broader academic and general convention, we use the former title to refer to the original work.}, containing 500 segments across diverse cultural categories.
Using a comprehensive evaluation protocol, we reveal three main challenges: 
(1) \textit{task challenges}, where frontier LLMs exhibit notable performance gaps and struggle with culturally loaded content; 
(2) \textit{human evaluation challenges}, where evaluator backgrounds lead to substantial disagreement in translation judgments; and 
(3) \textit{automatic evaluation challenges}, where widely used metrics fail to reliably assess translation quality for this task.
These findings may offer valuable insights for culture-oriented translation research in both computational science and linguistics.
\end{abstract}

\section{Introduction}

\begin{figure*}[t]
\vspace{-0.3in}
  \centering
  \includegraphics[width=2.05\columnwidth]{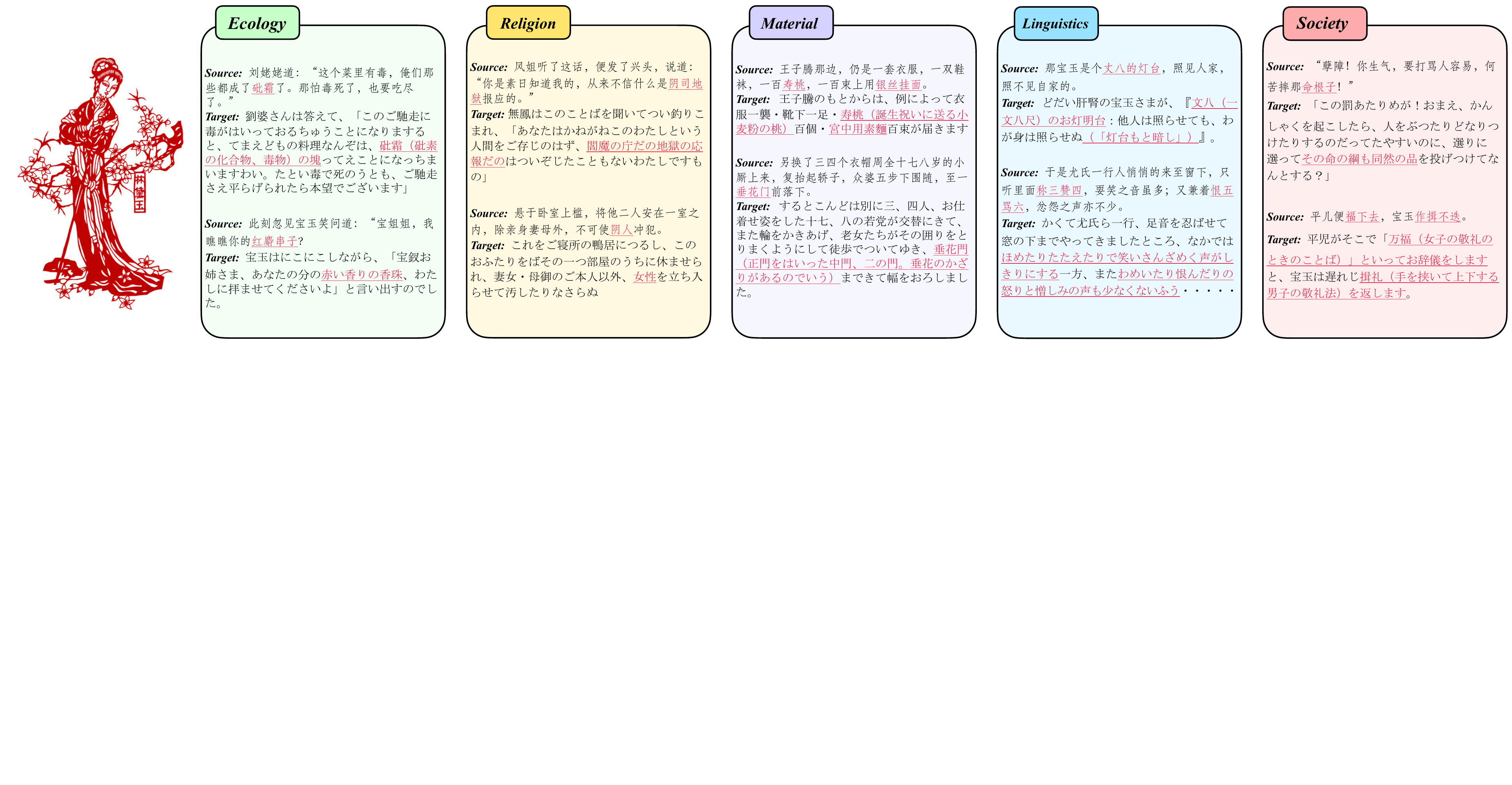} 
  \vspace{-1.9in}
  \caption{Dataset examples from our \textit{Dream of the Red Chamber} corpus across the five cultural categories. Detailed category definitions are in \textcolor{deepred}{§\ref{sec:category}}, and more examples are provided in \textcolor{deepred}{§\ref{appe:examples}}.}
  \label{fig:main-cases}
\vspace{-0.2in}
\end{figure*}

Translation aims to convey the source message in the target language through the closest natural equivalent while preserving semantic, stylistic, and pragmatic effects. However, language is inseparable from culture, and cultural priorities shape linguistic expression \citep{nida1964toward}. Each culture organizes its vocabulary around its own areas of emphasis, leading certain domains of meaning to become increasingly detailed and complex \citep{larson1997meaning}.
This culturally shaped development gives rise to a category of lexical items termed {\it culturally loaded words} by \citet{xu1980culturally}. Such words are deeply embedded in specific socio-cultural contexts and reflect the traditions, beliefs, and value systems of a community \citep{baker1992other,newmark1998more}.

The translation of culturally loaded expressions is {\bf highly contextualized}. Its context extends beyond the immediate text to the broader cultural background behind corresponding languages. Differences in cultural focus across languages may lead to semantic mismatches or lexical gaps \citep{newmark1988textbook,bader1994mona}. Therefore, translators must consider both source and target cultural contexts and balance interpretation with communication.
{\it At the interpretive level}, they should pursue conceptual equivalence rather than rely on surface correspondence, ensuring that target readers grasp the intended meaning, since literal translation is often insufficient \citep{nida1964toward}. 
{\it At the communicative level}, these expressions carry substantial cultural weight and should be rendered in ways that preserve, as far as possible, meanings rooted in the source culture \citep{venuti1995translator}. 
These place high demands on the professional expertise of translators.

On the computer science side, the rapid development of large language models (LLMs) \citep{brown2020language,touvron2023llama,bai2023qwen} has enabled machine translation (MT) systems to achieve human-like quality in many scenarios \citep{vilar2023prompting,jiao2023chatgpt,he2024exploring}. However, capturing rich contextual information remains a major challenge for current MT systems \citep{vieira2021understanding,moghe2025machine,zhang2025good,shen2025liaozhai}, making the culturally loaded translation task particularly difficult for current models due to its inherently contextualized nature.
Evaluating such translations adds another layer of difficulty. Translation quality is already subjective, and cultural elements amplify this subjectivity: evaluators from different cultural backgrounds may judge the same output differently based on their cultural familiarity and preferences (see \textcolor{deepred}{§\ref{sec:theory}} for the theoretical basis), introducing unpredictable variance in human evaluation. Furthermore, it remains unclear whether widely used automatic metrics can reliably assess translation quality in these scenarios.
Thus, from the task itself to its evaluation, culturally loaded translation presents unique challenges for modern MT research.

In this study, we systematically investigate the above challenges that it poses for LLM-based MT.
We construct a Chinese-Japanese bilingual dataset (\textcolor{deepred}{§\ref{sec:corpus}}) based on the highly representative cultural corpus of {\it Dream of the Red Chamber}. The dataset comprises 500 representative culturally loaded segments, and spans diverse cultural categories, with representative examples shown in \textcolor{deepred}{Fig.\ref{fig:main-cases}}.
We also design a specific evaluation protocol (\textcolor{deepred}{§\ref{sec:protocol}}) that incorporates diverse evaluation dimensions and evaluator backgrounds, ensuring the completeness and comprehensiveness of the results.
All preliminary model translation qualities are based on human evaluation, which serves as the gold standard for our formal analysis.
Based on these, we reveal three comprehensive core challenges:

\vspace{-0.02in}
\begin{itemize}[leftmargin=0.1cm]
    \item \textbf{Task Difficulty (\textcolor{deepred}{§\ref{sec:model-analysis}}):} Frontier LLMs still underperform on culturally loaded translation, showing substantial gaps both across models and relative to human references. Their performance also varies across cultural categories and is sensitive to category-specific textual characteristics and difficulty. This indicates considerable room for improvement in LLM-based MT systems.
    \vspace{-0.02in}
    \item \textbf{Human Evaluation Disagreement (\textcolor{deepred}{§\ref{sec:evaluator-analysis}}):} Human evaluation of culturally loaded translation is sensitive to evaluator background, including native cultural environment and expertise. This indicates that evaluating this task requires careful design of evaluator diversity and comprehensiveness; otherwise, systematic bias may arise.
    \vspace{-0.02in}
    \item \textbf{Automatic Evaluation Unreliability (\textcolor{deepred}{§\ref{sec:sample-corr}}):} Mainstream automatic metrics struggle to reliably capture model rankings and totally fail to distinguish sample-level quality differences in culturally loaded translation. This highlights the need for more task-aware automatic methods.
\end{itemize}
\vspace{-0.02in}

In addition, we conduct extended error and translation strategy analyses to better understand model behavior patterns and key bottlenecks in this task. We hope these findings provide valuable insights for both linguistic and computational research.

\section{Theoretical Foundations}
\label{sec:foundations}

\subsection{Culturally Loaded Translation}

\citet{xu1980culturally} first coined the term \textit{culturally loaded words} to describe lexical items that encode society's traditions, beliefs, and value systems. Similar concepts in translation studies, such as \textit{cultural words} \citep{newmark1988textbook} and \textit{culture-specific items} \citep{aixela1996culture}, generally refer to expressions that are deeply rooted in particular socio-cultural contexts and whose meanings cannot be fully understood without cultural background knowledge \citep{baker1992other,newmark1998more}.

Taking Chinese culture as examples, \begin{CJK}{UTF8}{gbsn}“布衣”\end{CJK} ({\it bùyī}) in ancient China referred to ordinary people and implied a modest lifestyle associated with coarse cloth garments. If translated literally as “coarse clothes,” English readers may find it confusing, since the phrase only refers to a type of fabric and does not convey the social meaning in Chinese.
Another example is \begin{CJK}{UTF8}{gbsn}“鸿雁”\end{CJK} ({\it hóngyàn}), a bird image in Chinese ecological culture. Although “swan goose” is its literal zoological equivalent, it fails to capture the cultural meaning. In Chinese tradition, \begin{CJK}{UTF8}{gbsn}“鸿雁”\end{CJK} often symbolizes a messenger carrying letters. Translating it as “message-bearing swans”\footnote{Translated by Chinese translator Yuanchong Xu.} better conveys this cultural imagery, even though it sacrifices strict lexical equivalence.
These examples show that translating culturally loaded expressions requires conveying both meaning and cultural significance.

\subsection{Cultural Category}
\label{sec:category}

As the foundational scholar of modern translation studies, Eugene A. Nida pioneered the categorization of culture into five types: ecology, religion, material, linguistics, and society \citep{nida1945linguistics,nida1964toward}:

\vspace{-0.06in}
\begin{itemize}[leftmargin=0.4cm]
\item \textbf{Ecology}: Natural elements such as flora, fauna, climate, geographical landscapes, and ecological phenomena that carry cultural meaning.
\vspace{-0.08in}
\item \textbf{Religion}: Spiritual beliefs including ancestor worship, folk superstitions, and mythological concepts that shape worldviews.
\vspace{-0.08in}
\item \textbf{Material}: Tangible artifacts of daily life like clothing, food, tools, and other physical objects that define a civilization's material culture.
\vspace{-0.08in}
\item \textbf{Linguistics}: Culturally embedded expressions like idioms, proverbs, slang, riddles, and fixed phrases that exhibit unique rhetorical patterns.
\vspace{-0.08in}
\item \textbf{Society}: Social structures, customs, institutions, kinship systems, etiquette norms, and conventions that govern interpersonal relations.
\end{itemize}
\vspace{-0.03in}

This classification has shown strong theoretical adaptability and continues to inform contemporary translation research \citep{oo2020culture,Yuwei2022EnglishTO,prompan2024study,al2025cultural}. In this study, we adopt this five-fold taxonomy as the framework for dataset construction.

\subsection{Cross-culture Translation Theory}
\label{sec:theory}

Cross-cultural translation has long been a central topic across disciplines. Among the most influential frameworks is Venuti's domestication and foreignization taxonomy \citep{venuti1995translator}. Domestication adapts a text to target-culture norms to enhance readability, whereas foreignization preserves source-culture elements to foreground cultural difference. This framework highlights the translator's agency and offers a critical perspective for evaluating cross-cultural transfer. It has been widely used to analyze how translators handle culture-specific items by balancing accessibility and authenticity.

Other theoretical perspectives further deepen the understanding of translation as a culturally embedded practice. The dynamic equivalence theory \citep{nida1964toward} prioritizes equivalent reader response rather than literal correspondence. The cultural turn thoery \citep{Bassnett1990TranslationHA} conceptualizes translation as a form of cultural rewriting shaped by ideological forces; The norms theory \citep{Toury1995DescriptiveTS} emphasizes the socio-cultural constraints that guide translators' decisions.
Together, these frameworks suggest that translation is not a neutral linguistic transfer but a process of cultural negotiation --- an important foundation for examining how language models handle culturally loaded translation, which we will also explore in our subsequent experiments.

\section{Corpus}
\label{sec:corpus}

\subsection{Data Source}
\label{sec:data-source}

We used \textit{Dream of the Red Chamber} as our source corpus. 
First, widely regarded as the ``encyclopedia of Chinese classical culture'' \citep{Edwards2005FictionsOE,plaks2015archetype}, it displays exceptional linguistic richness and cultural depth. 
Second, it covers all five cultural categories in \textcolor{deepred}{§\ref{sec:category}}, enabling comprehensive evaluation across cultural dimensions within a unified textual framework \citep{Xia2011OnET}. 
Third, its long translation history and the extensive scholarship surrounding it, often referred to as ``redology'' \citep{saussy1995problem,moratto2022dream}, allow findings derived from this corpus to connect with broader research and cross-cultural communication.

\subsection{Target Language}
\label{sec:target-lang}

Building on Chinese culture as the source, we select Japanese as the target language, as its cultural interpretive distance \citep{Bassnett1990TranslationHA,Toury1995DescriptiveTS} from Chinese strikes a balance.
Unlike translations between Chinese and Western languages, which often require extensive cultural explanation, Chinese and Japanese belong to the East Asian cultural sphere and share elements such as the kanji writing system and related aesthetic traditions. This shared foundation allows many culturally loaded concepts to be transferred more directly, while still requiring careful strategic choices when meanings diverge.
For example, the Chinese term ``\begin{CJK}{UTF8}{gbsn}阴司\end{CJK}'' ({\it yīnsī}) is rendered in the Japanese translation as \begin{CJK}{UTF8}{min}「閻魔の庁」\end{CJK}. This substitution draws on shared religious conceptions between Chinese and Japanese cultures, where both traditions envision judgment by King Yama after death. In contrast, Western readers' imagination of the ``nether world'' tends to evoke Greek Hades or the Christian purgatory, lacking the distinct East Asian imagery of Yama's judgment.
This intermediate cultural distance makes the Chinese–Japanese pair particularly revealing for studying culturally loaded translation.

\begin{table}[t]
\caption{Mean lengths and standard deviations for source and target texts per cultural category.}
\vspace{-0.1in}
\centering
\footnotesize
\renewcommand\arraystretch{1.1}
\setlength{\tabcolsep}{1.8mm}{
\resizebox{1\columnwidth}{!}{
\begin{tabular}{l|ccccc}
\toprule
& {\bf Ecology} & {\bf Religion} & {\bf Material} & {\bf Language} & {\bf Society} \\
\midrule
{\bf Source} & 
23.40\textsubscript{±12.35} & 
31.70\textsubscript{±16.21} & 
24.20\textsubscript{±16.69} & 
15.20\textsubscript{±9.42} & 
24.90\textsubscript{±24.19} \\

{\bf Target} & 
50.00\textsubscript{±25.43} & 
64.30\textsubscript{±36.92} & 
56.50\textsubscript{±44.72} & 
32.05\textsubscript{±19.42} & 
51.05\textsubscript{±43.33} \\
\bottomrule
\end{tabular}}}
\label{tab:length}
\vspace{-0.2in}
\end{table}

\subsection{Translation Version}
\label{sec:trans-version}

We prioritize target-native translations because conveying culturally loaded meanings in the target language requires generative competence in its linguistic and cultural norms \citep{nida1993language,bassnett2007culture}. Mastery of the target culture is thus more consequential than source-culture familiarity, leading us to adopt a translation by a native Japanese speaker.

Japanese translations of {\it Dream of the Red Chamber} date back to Mori Kainan in 1892. Over the following century, they evolved from partial renderings into at least 38 full versions. Among these, the full translations by Matsueda Shigeo, Ito Sohei, and Inami Ryoichi are widely regarded as the three pillars of the Japanese tradition \citep{WJYY201901011}. We adopt Ito Sohei's version for this study. Unlike Inami Ryoichi's modern Japanese rendering, which prioritizes reader accessibility, Ito's translation achieves a more balanced approach by preserving source fidelity while providing rich annotations that deepen cultural interpretation \citep{wu2019yanshi}. Furthermore, developed over nearly fifty years with four revisions, it has become the most influential and widely studied version in academic research \citep{Hu1993, wan1998, sun2006}.

\subsection{Collection and Annotation}
\label{sec:annotation}

We collected the bilingual Chinese-Japanese edition from the state-sponsored \textit{Library of Chinese Classics} series\footnote{https://zh.wikipedia.org/wiki/\begin{CJK}{UTF8}{gbsn}大中华文库\end{CJK}}, translated by Ito Sohei. This edition was compiled under the auspices of Chinese government agencies and extensively revised by renowned scholars, ensuring its authority and reliability. Also, its format enables direct comparisons between source and target texts (see \textcolor{deepred}{§\ref{appe:pdf}}), ensuring rigor throughout the data processing.

Currently, only image versions of this edition are available, so OCR processing is required. Four Chinese graduate students conducted the initial annotation, each selecting 480 culturally loaded segments (4 per chapter across all 120 chapters) and categorizing them according to the taxonomy in \textcolor{deepred}{§\ref{sec:category}}. Four Japanese graduate students then performed a second round of screening and revision. To control the workload of subsequent human evaluation, the dataset was further refined to 500 representative segments, balanced across the five cultural categories (100 per category).
The entire annotation process took 20 days. These annotators also participated in human evaluations, so their backgrounds are presented in subsequent \textcolor{deepred}{§\ref{sec:human-protocol}}.

Text length statistics are shown in \textcolor{deepred}{Tab.\ref{tab:length}} using the Qwen3 tokenizer \citep{yang2025qwen3}.
Overall, target texts are about twice as long as source texts. Among categories, {\it Religion} tends to be longer, {\it Linguistics} shorter, and the other three are similar.
Importantly, there are no marked differences, meaning that category difficulty is not driven by text length.

\section{Evaluation Protocol}
\label{sec:protocol}

\begin{table}[t]
\caption{Language models used in our experiments.}
\vspace{-0.1in}
\centering
\footnotesize
\renewcommand\arraystretch{1.1}
\setlength{\tabcolsep}{1.8mm}{
\resizebox{1\columnwidth}{!}{
\begin{tabular}{l|lll}

\toprule
{\bf Model Name} & {\bf Abbreviation} & {\bf Affiliation} & {\bf Citation} \\
\midrule

\multicolumn{4}{c}{Non-Reasoning Model} \\
\midrule

Deepseek-v3 & {\it DS-v3} & Deepseek & \citet{liu2024deepseek} \\
Qwen3-235B-A22-Non-Thinking & {\it Qwen3-T} & Alibaba & \citet{yang2025qwen3} \\
GPT-4.1 & {\it GPT-4.1} & OpenAI & \citet{gpt41} \\
Gemini-2.5-Flash & {\it Gemini-2.5} & Google & \citet{comanici2025gemini} \\

\midrule
\multicolumn{4}{c}{Reasoning Model} \\
\midrule

Deepseek-r1 & {\it DS-r1} & Deepseek & \citet{guo2025deepseek} \\
Qwen3-235B-A22-Thinking & {\it Qwen3-NT} & Alibaba & \citet{yang2025qwen3} \\
OpenAI-o4-mini & {\it o4-mini} & OpenAI & \citet{o4mini} \\
Claude-Sonnet-4 & {\it Claude-4} & Anthropic & \citet{claude4} \\

\bottomrule

\end{tabular}}}
\label{tab:model}
\vspace{-0.1in}
\end{table}

\begin{table}[t]
\caption{Evaluation dimensions used in our experiments.}
\vspace{-0.1in}
\centering
\footnotesize
\renewcommand\arraystretch{1.1}
\setlength{\tabcolsep}{2.5mm}{
\resizebox{1\columnwidth}{!}{
\begin{tabular}{l|lll}
\toprule
\textbf{Dimension Name} & \textbf{Abbreviation} & \textbf{Target} & \textbf{Scope} \\
\midrule
Content Accuracy & \textit{Acc.} & Content & General \\
Language Fluency & \textit{Flu.} & Language & General \\
Cultural Appropriateness & \textit{Cult.} & Content & Culture-specific \\
Native Readability & \textit{Read.} & Language & Culture-specific \\
\bottomrule
\end{tabular}}}
\label{tab:metric}
\vspace{-0.2in}
\end{table}

\begin{table*}[t]
\vspace{-0.1in}
\centering
\caption{Human evaluation results for eight models and one reference across five cultural categories and the full dataset. For each category, we report overall quality scores ({\it Overall}) along with four evaluation dimensions. \gold{}, \silver{}, and \bronze{} mark the top three performing models in each dimension.}
\vspace{-0.1in}
\resizebox{\textwidth}{!}{
\begin{tabular}{lccccc|ccccc|ccccc}
\toprule
& \multicolumn{5}{c}{\bf Full Dataset} 
& \multicolumn{5}{c}{\bf Ecology} 
& \multicolumn{5}{c}{\bf Religion} \\
\cmidrule(lr){2-6} \cmidrule(lr){7-11} \cmidrule(lr){12-16}
{\bf Model Name} & {\it Overall} & {\it Acc.} & {\it Flu.} & {\it Cult.} & {\it Read.} 
      & Overall & {\it Acc.} & {\it Flu.} & {\it Cult.} & {\it Read.}
      & Overall & {\it Acc.} & {\it Flu.} & {\it Cult.} & {\it Read.} \\
\midrule

{\it Reference} 
& {\it 4.27} & {\it 4.35} & {\it 4.24} & {\it 4.28} & {\it 4.11} 
& {\it 4.33} & {\it 4.38} & {\it 4.27} & {\it 4.32} & {\it 4.16} 
& {\it 4.23} & {\it 4.33} & {\it 4.29} & {\it 4.24} & {\it 4.09}
\\

Deepseek-v3
& \bronze{3.83} & 3.92 & \silver{4.05} & \gold{3.83} & 3.67
& \silver{4.06} & \silver{4.11} & \silver{4.24} & \gold{4.05} & \bronze{3.89}
& \silver{3.88} & 3.92 & \silver{4.15} & \gold{3.83} & 3.67 \\

Qwen3-235B-A22B-Non-Thinking
& 3.30 & 3.60 & 3.46 & 3.37 & 3.40
& 3.49 & 3.79 & 3.66 & 3.59 & 3.56
& 3.33 & 3.71 & 3.54 & 3.42 & 3.49 \\

GPT-4.1
& 3.66 & 3.75 & 3.72 & 3.63 & \gold{3.94}
& 3.82 & 3.89 & 3.85 & 3.71 & \gold{4.09}
& 3.64 & 3.84 & 3.71 & \bronze{3.61} & \gold{3.91} \\

Gemini-2.5-Flash
& \silver{3.87} & \silver{3.94} & \bronze{3.91} & \silver{3.82} & \silver{3.90}
& \bronze{4.03} & \bronze{4.10} & \bronze{4.00} & \silver{3.90} & \silver{4.05}
& \bronze{3.84} & \silver{3.99} & \bronze{3.89} & \silver{3.74} & \silver{3.88} \\

Deepseek-r1
& 3.37 & 3.51 & 3.53 & 3.33 & 3.48
& 3.51 & 3.60 & 3.61 & 3.38 & 3.60
& 3.34 & 3.56 & 3.49 & 3.25 & 3.47 \\

Qwen3-235B-A22-Thinking
& 3.65 & \bronze{3.93} & 3.76 & 3.63 & 3.31
& 3.85 & \silver{4.11} & 3.88 & 3.81 & 3.50
& 3.65 & \bronze{3.96} & 3.81 & \bronze{3.61} & 3.38 \\

OpenAI-o4-mini
& \gold{3.89} & \gold{4.00} & \gold{4.17} & \bronze{3.74} & \bronze{3.72}
& \gold{4.09} & \gold{4.20} & \gold{4.33} & \bronze{3.89} & \bronze{3.89}
& \gold{3.90} & \gold{4.06} & \gold{4.23} & \silver{3.74} & \bronze{3.78} \\

Claude-Sonnet-4
& 3.57 & 3.78 & 3.80 & 3.64 & 3.42
& 3.73 & 3.91 & 3.92 & 3.72 & 3.58
& 3.54 & 3.85 & 3.78 & \bronze{3.61} & 3.39 \\

\midrule
\textbf{Average}
& \textbf{3.64} & \textbf{3.80} & \textbf{3.80} & \textbf{3.62} & \textbf{3.61}
& \textbf{3.82} & \textbf{3.96} & \textbf{3.94} & \textbf{3.76} & \textbf{3.77}
& \textbf{3.64} & \textbf{3.86} & \textbf{3.83} & \textbf{3.60} & \textbf{3.62} \\

\textbf{Range (max-min)}
& \textbf{0.59} & \textbf{0.49} & \textbf{0.71} & \textbf{0.50} & \textbf{0.63}
& \textbf{0.60} & \textbf{0.60} & \textbf{0.72} & \textbf{0.67} & \textbf{0.59}
& \textbf{0.57} & \textbf{0.50} & \textbf{0.74} & \textbf{0.58} & \textbf{0.53} \\
\midrule

& \multicolumn{5}{c}{\bf Material} 
& \multicolumn{5}{c}{\bf Linguistics} 
& \multicolumn{5}{c}{\bf Society} \\
\cmidrule(lr){2-6} \cmidrule(lr){7-11} \cmidrule(lr){12-16}

{\bf Model Name} & {\it Overall} & {\it Acc.} & {\it Flu.} & {\it Cult.} & {\it Read.}
      & Overall & {\it Acc.} & {\it Flu.} & {\it Cult.} & {\it Read.}
      & Overall & {\it Acc.} & {\it Flu.} & {\it Cult.} & {\it Read.} \\

\midrule

{\it Reference} 
& {\it 4.39} & {\it 4.39} & {\it 4.33} & {\it 4.34} & {\it 4.17} 
& {\it 4.17} & {\it 4.29} & {\it 4.16} & {\it 4.22} & {\it 4.05} 
& {\it 4.23} & {\it 4.36} & {\it 4.15} & {\it 4.28} & {\it 4.08} 
\\

Deepseek-v3
& \silver{4.08} & 4.12 & \silver{4.26} & \gold{4.09} & 3.88
& \bronze{3.59} & 3.73 & \silver{3.86} & \silver{3.61} & 3.51
& \bronze{3.54} & \bronze{3.68} & \silver{3.75} & \silver{3.58} & 3.42 \\

Qwen3-235B-A22B-Non-Thinking
& 3.56 & 3.84 & 3.72 & 3.67 & 3.64
& 3.10 & 3.45 & 3.33 & 3.20 & 3.28
& 3.00 & 3.35 & 3.15 & 3.10 & 3.14 \\

GPT-4.1
& 3.85 & 3.93 & 3.86 & 3.72 & \gold{4.13}
& 3.53 & 3.66 & 3.65 & 3.52 & \gold{3.83}
& 3.45 & 3.58 & 3.50 & 3.45 & \gold{3.76} \\

Gemini-2.5-Flash
& \silver{4.08} & \bronze{4.15} & \bronze{4.04} & \bronze{3.94} & \silver{4.10}
& \gold{3.71} & \silver{3.79} & \bronze{3.81} & \gold{3.67} & \silver{3.74}
& \gold{3.68} & \gold{3.76} & \bronze{3.71} & \gold{3.65} & \silver{3.72} \\

Deepseek-r1
& 3.59 & 3.68 & 3.69 & 3.46 & 3.70
& 3.22 & 3.38 & 3.43 & 3.20 & 3.35
& 3.18 & 3.34 & 3.33 & 3.18 & 3.33 \\

Qwen3-235B-A22-Thinking
& \bronze{3.90} & \silver{4.17} & 3.92 & 3.85 & 3.56
& 3.47 & \bronze{3.75} & 3.64 & 3.49 & 3.21
& 3.37 & 3.65 & 3.47 & 3.40 & 3.10 \\

OpenAI-o4-mini
& \gold{4.13} & \gold{4.19} & \gold{4.40} & \silver{3.96} & \bronze{3.97}
& \silver{3.69} & \gold{3.81} & \gold{4.00} & \bronze{3.58} & \bronze{3.57}
& \silver{3.62} & \silver{3.74} & \gold{3.88} & \bronze{3.52} & \bronze{3.49} \\

Claude-Sonnet-4
& 3.79 & 3.98 & 3.97 & 3.77 & 3.64
& 3.43 & 3.66 & 3.71 & 3.51 & 3.32
& 3.38 & 3.61 & 3.60 & 3.48 & 3.27 \\

\midrule
\textbf{Average}
& \textbf{3.87} & \textbf{4.01} & \textbf{3.98} & \textbf{3.81} & \textbf{3.83}
& \textbf{3.47} & \textbf{3.65} & \textbf{3.68} & \textbf{3.47} & \textbf{3.48}
& \textbf{3.40} & \textbf{3.59} & \textbf{3.55} & \textbf{3.42} & \textbf{3.40} \\

\textbf{Range (max-min)}
& \textbf{0.57} & \textbf{0.51} & \textbf{0.71} & \textbf{0.63} & \textbf{0.57}
& \textbf{0.61} & \textbf{0.43} & \textbf{0.67} & \textbf{0.47} & \textbf{0.62}
& \textbf{0.68} & \textbf{0.42} & \textbf{0.73} & \textbf{0.55} & \textbf{0.66} \\
\bottomrule
\end{tabular}}
\label{tab:human-eval}
\vspace{-0.15in}
\end{table*}

We evaluated eight language models (see \textcolor{deepred}{Tab.\ref{tab:model}}), selected for two aspects of diversity: (1) they span five organizations and multiple model families, providing broad source coverage; (2) they include both reasoning and non-reasoning paradigms, enabling us to examine the impact of long chain-of-thought (CoT) ability \citep{wei2022chain,guo2025deepseek} on culturally loaded translation.
Implementation details are provided in \textcolor{deepred}{§\ref{appe:implementation}}. Based on model outputs, we established the following evaluation protocol.

\subsection{Human Evaluation}
\label{sec:human-protocol}

Given the contextual complexity and inherent subjectivity of this task, our evaluation is human-centered, with human judgments serving as the gold standard.
We adopt the standard MQM framework \citep{lommel2013multidimensional}, a widely used functionalist evaluation approach\footnote{The standard MQM framework includes five dimensions: Accuracy, Fluency, Verity, Design, and Internationalization.}, and adapt it to better capture the characteristics of culturally loaded translation. Specifically, we define four evaluation dimensions, shown in \textcolor{deepred}{Tab.\ref{tab:metric}} (see \textcolor{deepred}{§\ref{appe:main}} for detailed explanations and scoring guidelines).
The first two assess semantic accuracy and linguistic fluency following MQM. The remaining two focus on cultural aspects, evaluating whether the translation conveys the cultural meaning of the source expression and whether it reads naturally to target-culture readers. In addition, evaluators assign an \textit{overall quality score} to reflect their holistic judgment.

Our evaluators come from diverse cultural and academic backgrounds and consist of sixteen native speakers, including eight Chinese and eight Japanese. Within each language group, four are graduate students specializing in the other language’s literature, and four are senior academics serving as university professors in the other country's cultural studies.
To ensure evaluation reliability, all student evaluators hold at least JLPT-N1 or HSK-6 certification, or an equivalent qualification.
This design yields four groups by language ({\it zh} / {\it ja}) and academic level ({\it stu.} / {\it prof.}).

Given the data scale, evaluating all translations would be prohibitively labor-intensive: (8 models + 1 reference) × 500 samples = 4,500 translations, resulting in 22,500 scores per evaluator.
Therefore, within each group, four evaluators each review 125 samples (25 per category), collectively covering the full dataset. For \textit{Read.}, scores are assigned only by target-language natives.
To avoid bias from prior assumptions about model and human abilities, translation sources were anonymized. Evaluators saw only the source text and nine shuffled translations for each sample.
Before the evaluation, pilot assessments verified intra-group agreement. Sampling-based comparisons showed that evaluators within the same group had comparable expertise and low judgment variance.
After the evaluation, we conducted follow-up interviews with the evaluators to examine their fine-grained decision criteria and interpretations of some anomalous phenomena, which are needed for analysis in \textcolor{deepred}{§\ref{sec:evaluator-analysis}}.

\subsection{Automatic Evaluation}

We adopt three types of metrics: lexical-based BLEU \citep{papineni2002bleu}, semantic-based xCOMET \citep{he2024exploring}, and LLM-as-a-Judge \citep{zheng2023judging}.
For the latter, we employ two models: GPT-5 (a non-reasoning model, \citet{singh2025openai}) and Gemini-3-Pro (a reasoning model, \citet{gemini3pro}).
See \textcolor{deepred}{§\ref{appe:llm-judge}} for the specific prompt.

\begin{figure*}[t]
\vspace{-0.1in}
  \centering
  \includegraphics[width=2\columnwidth]{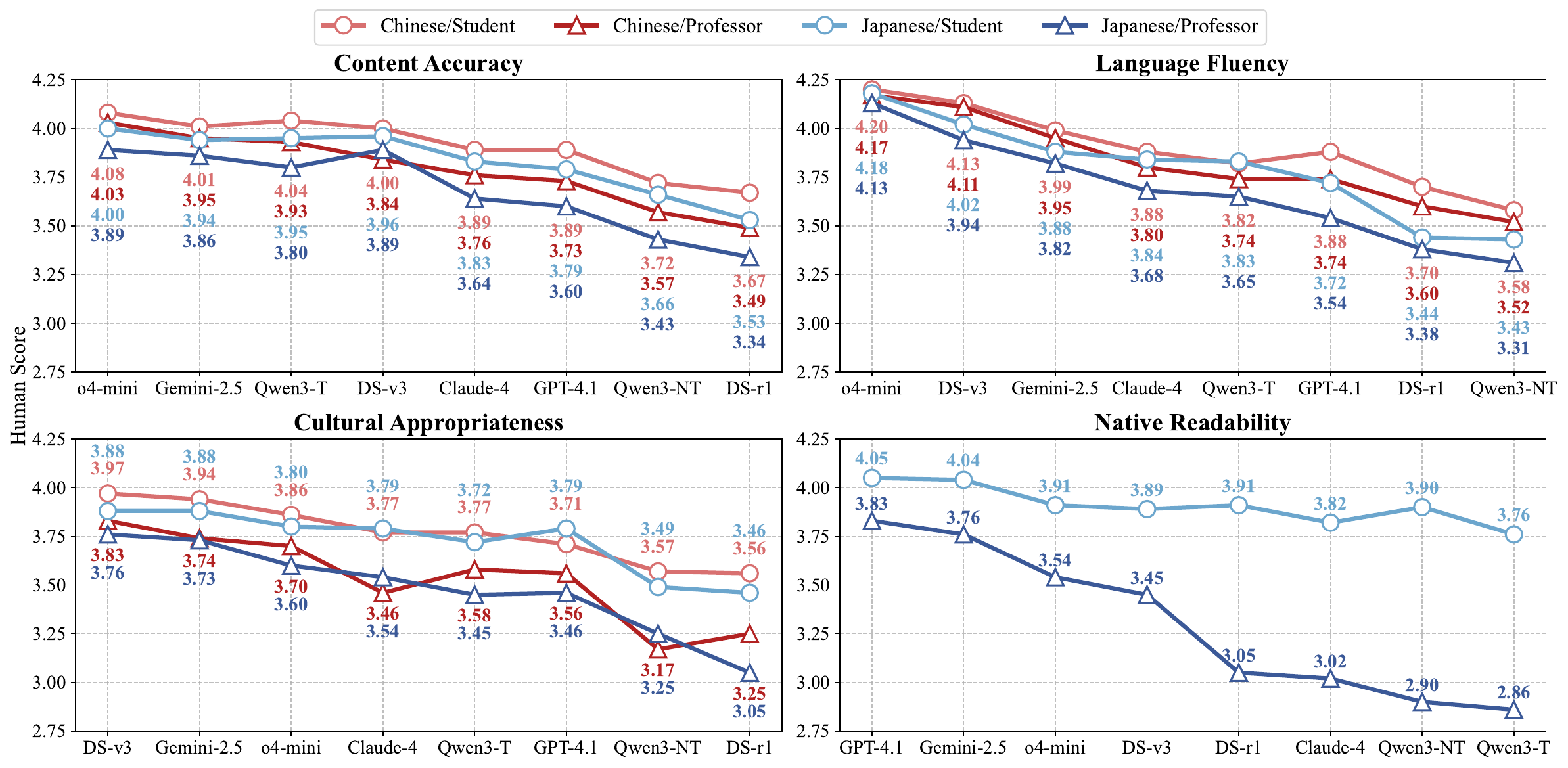}
  \vspace{-0.1in}
  \caption{Human evaluation results for eight models across four dimensions, scored by four evaluator groups: Chinese/Student ({\it zh-stu.}), Chinese/Professor ({\it zh-prof.}), Japanese/Student ({\it ja-stu.}), Japanese/Professor ({\it ja-prof.}). In each subfigure, {\it models are arranged from left to right in descending order of their overall scores on the full dataset}.}
  \label{fig:meta}
\vspace{-0.2in}
\end{figure*}

\section{Results}

\subsection{Task Difficulty (\textcolor{deepred}{Tab.\ref{tab:human-eval}})}
\label{sec:model-analysis}

\paragraph{Model Performance Comparison.}
We first analyze overall model performances on the full dataset.
The top three models in overall score are \textit{OpenAI-o4-mini} (3.89), \textit{Gemini-2.5-Flash} (3.87), and \textit{Deepseek-v3} (3.83). Notably, the first two rank among the top three across all five evaluation dimensions, indicating strong robustness.
However, performances vary greatly across models, with overall scores ranging from 3.89 to 3.30. This wide range demonstrates that this task poses a meaningful challenge and effectively differentiates model capabilities.
In addition, a considerable gap remains between current models (an average of 3.64) and human references (4.27), whose scores are consistently higher across all dimensions, indicating that culturally loaded translation remains difficult for current LLMs to match human performance.

Another interesting phenomenon is that reasoning and non-reasoning models reveal no systematic advantage for either group. Their average overall scores are nearly identical (3.66 vs. 3.62), and medal counts are similarly balanced (15 vs. 12 medals, with 2 vs. 3 golds). While the top-performing \textit{OpenAI-o4-mini} employs reasoning, the runner-ups \textit{Gemini-2.5-Flash} and \textit{Deepseek-v3} do not, suggesting that longer CoT are not decisive for our culture-oriented translation tasks.

\paragraph{Dimension Difference.}
Next, we analyze model performance across evaluation dimensions.
A clear pattern emerges: the two culturally related dimensions, \textit{Cult.} and \textit{Read.}, consistently receive lower scores than the general two, \textit{Acc.} and \textit{Flu.}.
This suggests that while current models perform relatively well on general quality, they struggle more with culturally loaded aspects of translation.
Moreover, strong performance on general dimensions does not necessarily imply strong cultural competence. For example, \textit{OpenAI-o4-mini} achieves the highest scores in \textit{Acc.}(4.00) and \textit{Flu.}(4.17), yet its \textit{Cult.}(3.74) and \textit{Read.}(3.72) only rank third. Conversely, \textit{GPT-4.1} attains the highest \textit{Read.} (3.94) but performs less strongly in general dimensions.
These phenomena indicate that cultural competence cannot be inferred directly from general translation quality, revealing a gap between linguistic correctness and deeper cultural adaptation.

\paragraph{Cultural Category Difference.}
\label{sec:category-analysis}

Finally, we analyze model performance across cultural categories. The results reveal a clear difficulty hierarchy: \textit{Material} (3.87) and \textit{Ecology} (3.82) score the highest, followed by \textit{Religion} (3.64), while \textit{Linguistics} (3.47) and \textit{Society} (3.40) lag significantly behind.

This gap is largely driven by differences in textual characteristics across cultural categories.
\textit{Material} and \textit{Ecology} texts mainly describe concrete objects, entities, and natural phenomena, which are relatively straightforward for models to translate. \textit{Religion} texts, although conceptually abstract, often contain narrative elements and real-world descriptions that models can partially handle using general knowledge.
In contrast, \textit{Linguistics} and \textit{Society} texts are considerably more challenging. \textit{Linguistics} frequently includes classical idioms, allusions, and elliptical expressions with dense meanings and flexible syntax, making accurate interpretation and translation difficult. \textit{Social Culture} involves etiquette and honorific expressions that depend on subtle social relationships, requiring pragmatic inference that models often fail to capture, leading to inappropriate tone or meaning.

\begin{figure*}[t]
\vspace{-0.1in}
  \centering
  \includegraphics[width=2\columnwidth]{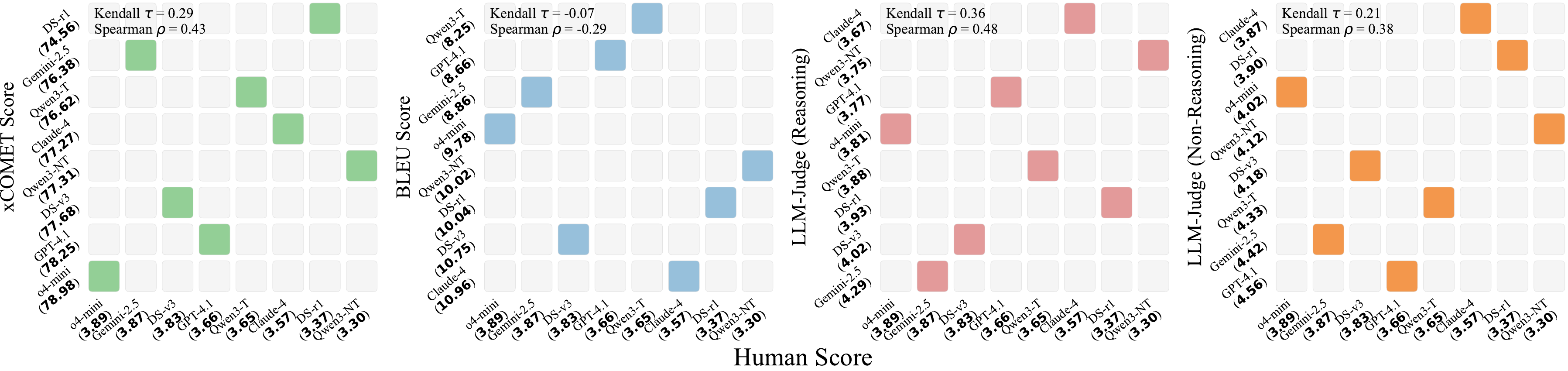} 
  \vspace{-0.12in}
  \caption{System-level correlation between human overall scores and automatic metrics. In each subplot, models on the $x$-axis are ordered by human scores from high to low, and models on the $y$-axis are ordered by the metric from low to high. Colored cells mark matching models, and Kendall’s $\tau$ and Spearman’s $\rho$ are shown in the top-left.}
  \label{fig:model-corr}
\vspace{-0.2in}
\end{figure*}

\subsection{Human Evaluation Disagreement (\textcolor{deepred}{Fig.\ref{fig:meta}})}
\label{sec:evaluator-analysis}

\paragraph{Overall Inter-group Comparison.}
The inter-group agreement varies across evaluation dimensions. For the first three dimensions, the differences among evaluator groups remain moderate, generally below 0.5 points. In contrast, the \textit{Read.} dimension shows substantially larger divergence. For several lower-ranked models, the gap between evaluator groups approaches 1 point. We analyze this anomaly in detail below.

Despite these variations, the relative ranking of the eight models remains largely consistent across evaluator groups and aligns well with the overall scores. Among the four groups, {\it ja-prof.} shows the closest agreement with the overall ranking, followed by {\it zh-prof.}. The two student groups exhibit slightly greater fluctuation, though the differences are not substantial. In addition, agreement tends to decrease as model quality declines, as poorer outputs leave more room for subjective interpretation.

\paragraph{Student vs. Professor.}

Within the same language group ({\it zh} or {\it ja}), students consistently score higher than professors. This may reflect differences in evaluation standards and familiarity with culturally loaded texts.
Through the post-evaluation interview, we find that when encountering ambiguous expressions, students tend to accept them more readily, which introduces an upward bias. By contrast, professors generally apply stricter criteria when judging translation adequacy.

This divergence is most evident in the \textit{Read.} dimension, where students and professors show substantial disagreement.
Through the post-evaluation interview, we note an important fact: culturally loaded texts are often associated with specific historical or literary registers. Students tend to evaluate readability primarily based on fluency and ease of comprehension. Professors, however, pay closer attention to whether the stylistic register (e.g. classical or modern language) appropriately reflects the cultural and historical context embedded in the source text, even if this reduces immediate readability.
In practice, models often produce relatively colloquial outputs that align with students' reading habits, leading them to score such translations more favorably. Professors, however, may consider these stylistically inappropriate, arguing that culturally significant texts should preserve more register-appropriate expressions.
This reflects a broader tension between accessibility and authenticity rather than a purely objective notion of correctness.

This is further evidenced by their scoring of the reference on the {\it Read.} dimension: students and professors gave average scores of 3.70 and 4.52, respectively.
Notably, the students' rating is even lower than that of some model outputs, providing additional support for the above conclusions.

\paragraph{Chinese (Source) vs. Japanese (Target).}

Within the same academic group ({\it stu.} or {\it prof.}), Japanese are slightly stricter than Chinese evaluators on the two general dimensions, \textit{Acc.} and \textit{Flu.}. This is expected, as Japanese annotators are native speakers of the target language and can therefore apply more nuanced judgments of translation quality.

In contrast, the relative scoring patterns for the cultural dimension \textit{Cult.} vary across models --- sometimes Japanese evaluators score higher, while in other cases Chinese evaluators do.
Through the post-evaluation interview, we find that although both groups value the preservation of source culture and the appropriateness of the target expression, they tend to emphasize different aspects. Chinese evaluators often focus more on faithfulness to the source culture, whereas Japanese evaluators place greater weight on appropriateness within the target cultural context.
Because current models often struggle to balance these two aspects simultaneously, translations that favor one side may draw stricter scrutiny from evaluators who prioritize the other, causing noticeable fluctuations in scoring.

\begin{figure*}[t]
\vspace{-0.1in}
  \centering
  \includegraphics[width=2\columnwidth]{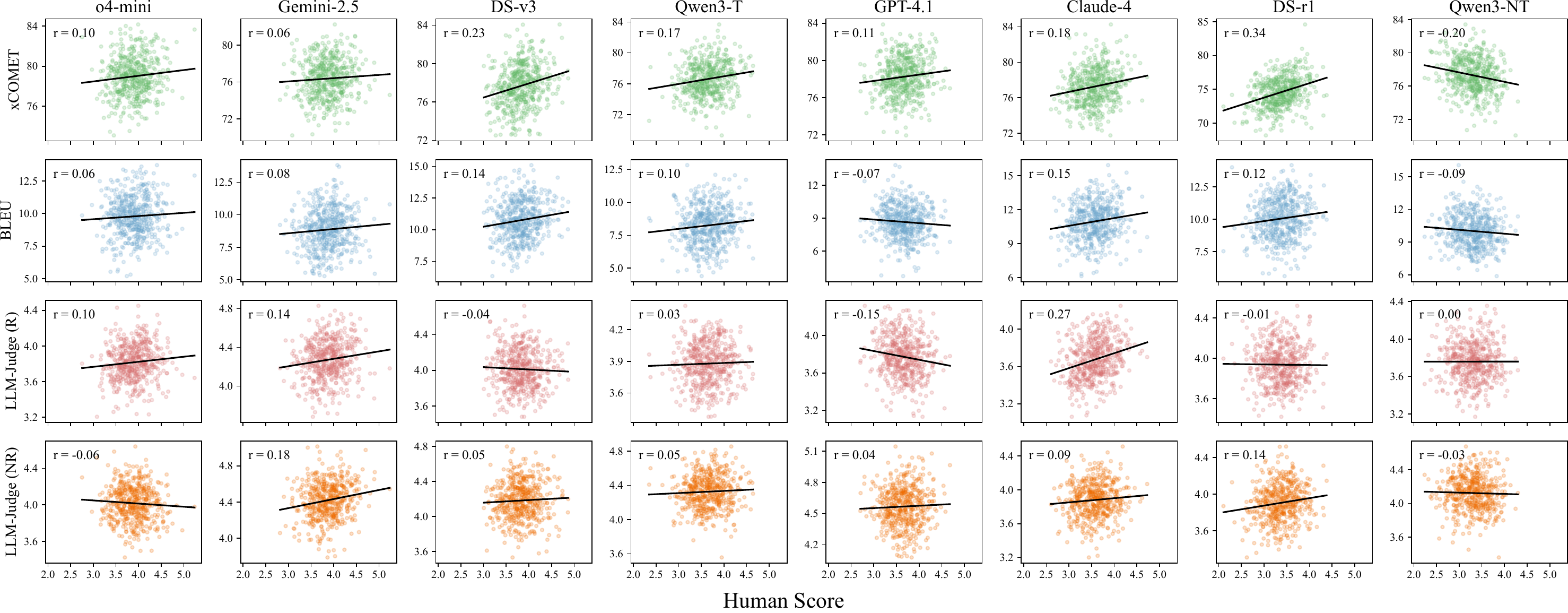}
  \vspace{-0.1in}
  \caption{Sample-level correlation between human overall scores and automatic metrics. Each subplot corresponds one model-metric pair with all 500 samples. The black line is the linear fit, and Pearson's $r$ are shown in the top-left.}
  \label{fig:sample-corr}
\vspace{-0.2in}
\end{figure*}

\subsection{Automatic Evaluation Unreliability (\textcolor{deepred}{Fig.}\ref{fig:model-corr}-\ref{fig:sample-corr})}
\label{sec:automatic-analysis}

\paragraph{System Correlation.}
\label{sec:system-corr}

\textcolor{deepred}{Fig.\ref{fig:model-corr}} compares system-level rankings between human evaluation and automatic metrics. Overall, correlations with human judgments remain limited. xCOMET and two LLM-based judges show modest positive correlations (Kendall's $\tau$ between 0.21 and 0.36, Spearman's $\rho$ between 0.38 and 0.48), with several models positioned near the diagonal. However, BLEU exhibits almost no correlation ($\tau<0.1$), as its reliance on surface-level lexical overlap with reference translations prevents it from capturing semantic and culturally grounded translation quality.
These results suggest that while newer metrics offer somewhat more informative signals than BLEU, they still fall short of reliably distinguishing model performance in culturally loaded translation tasks.

\paragraph{Sample Correlation.}
\label{sec:sample-corr}

\textcolor{deepred}{Fig.\ref{fig:sample-corr}} illustrates the correlation between human evaluation scores and four automatic metrics across all eight models. Overall, correlations with human judgments remain extremely weak at the sample level for all metrics. Although xCOMET and the LLM judges occasionally show slightly stronger positive correlations than BLEU, the overall alignment remains limited, and the scatter plots display substantial dispersion around the fitted regression lines.
In many cases, samples with similar automatic scores receive noticeably different human ratings, while samples with comparable human judgments correspond to a wide range of automatic metric values. This pattern suggests that automatic metrics capture only part of the translation quality signal and fail to reliably reflect fine-grained differences among individual outputs.
Consequently, the magnitude of an automatic metric score alone is insufficient to determine whether a specific generated sample is actually better or worse according to human evaluation.

\section{Extend Analysis}

To further explore the model limitations in translating culturally loaded content, we conduct extended analyses of their translation behavior.
Due to space limitations, detailed discussions are provided in the appendix, with key conclusions presented here.

\paragraph{Error Type (\textcolor{deepred}{§\ref{appe:error-analysis}}).}
We classify translation errors into five types: {\it No Obvious Error}, {\it Mistranslation}, {\it Overtranslation}, {\it Undertranslation}, and {\it Omission}. {\it Mistranslation} is the most frequent error across all models, while higher-performing models exhibit a lower proportion of such errors. The remaining three error types occur relatively infrequently. Notably, reasoning models produce substantially more {\it Overtranslation}, yet this tendency does not translate into better overall translation quality.

\paragraph{Translation Strategy (\textcolor{deepred}{§\ref{appe:strategy-analysis}}).}
We focus on the domestication-foreignization framework introduced in \textcolor{deepred}{§\ref{sec:theory}}. We find that over 80\% of translations adopt a single strategy, with a clear preference for domestication.
Under this, models tend to rigidly adapt cultural expressions, often sacrificing source cultural imagery for target-language readability, resulting in notable loss of cultural transmission.

\paragraph{Case Study (\textcolor{deepred}{§\ref{appe:case-study}}).}
To provide a more intuitive understanding of model translation styles, we present several case studies that illustrate how models underperform across various evaluation dimensions.

\section{Conclusion}

Culturally loaded translation has long been a central concern in translation studies, as language serves not only as a means of communication but also as a carrier of culture.
In this work, we connect this linguistic topic with modern MT research\footnote{In recent MT research, several culture-oriented translation studies have also emerged; we discuss them in \textcolor{deepred}{§\ref{appe:related-work}}.}. Using a dataset based on Chinese-Japanese translations of \textit{Dream of the Red Chamber}, we systematically uncover the challenges of culturally loaded translation from three perspectives: the task itself, human evaluation, and automatic evaluation.
Our analysis may offer insights for future research on culturally loaded translation in computer science.
More broadly, the results suggest that, despite rapid progress in LLMs, MT still has a long way to go, as many researchers have long believed \citep{kocmi2024findings,pang2025salute,ataman2025machine}.

\section*{Limitations}

The primary limitation of this study is language diversity. Although we have justified in \textcolor{deepred}{§\ref{sec:target-lang}} the representativeness of the current language pair for studying culturally loaded translation, including additional languages, particularly those with broader speaker populations such as English, would undoubtedly enhance the generalizability of our findings. However, due to various constraints, including limited funding, the lengthy cycle of human evaluation, and the need for culturally proficient native speakers (crowdsourcing is not a viable option for this study), we are currently unable to extend our study to more languages in the near term. As a mitigation, we will open-source our data sources and provide raw multilingual corpus to facilitate further research. We believe that future work can build upon this foundation to yield more valuable insights.

\section*{Ethical Considerations}

\paragraph{Potential Risk.}
Our study involves engineering experiments and analyses, and does not propose any new architectures or algorithms. Therefore, it does not introduce uncontrollable consequences in practical use.
Regarding the research data, we have carefully verified the compliance of both the constructed dataset and the translations generated by LLMs, ensuring no potential risks are involved.

\paragraph{Data Compliance.}
We have confirmed that our research data, including the constructed dataset and the translations generated by LLMs, does not contain any sensitive elements related to politics, violence, or similar topics.
We ensure that no uncontrollable impact will be caused to the human evaluators involved in the experiment or to any individuals who may come into contact with the data.

\paragraph{Human Evaluation.}
All evaluators were fully informed about the research use of the evaluation data collected from them, and their consent was obtained.
Throughout the evaluation process, we consistently respected their autonomy and paid them with mutually agreed-upon remuneration.

\section*{AI Assistants Declaration}

AI assistants (ChatGPT and Deepseek) were used exclusively for writing polishing and played no role in any other aspect of this work.


\bibliography{custom}

@article{guo2025deepseek,
  title={DeepSeek-R1 incentivizes reasoning in LLMs through reinforcement learning},
  author={Guo, Daya and Yang, Dejian and Zhang, Haowei and Song, Junxiao and Wang, Peiyi and Zhu, Qihao and Xu, Runxin and Zhang, Ruoyu and Ma, Shirong and Bi, Xiao and others},
  journal={Nature},
  volume={645},
  number={8081},
  pages={633--638},
  year={2025},
  publisher={Nature Publishing Group UK London}
}

@article{liu2024deepseek,
  title={Deepseek-v3 technical report},
  author={Liu, Aixin and Feng, Bei and Xue, Bing and Wang, Bingxuan and Wu, Bochao and Lu, Chengda and Zhao, Chenggang and Deng, Chengqi and Zhang, Chenyu and Ruan, Chong and others},
  journal={arXiv preprint arXiv:2412.19437},
  year={2024}
}

@article{yang2025qwen3,
  title={Qwen3 technical report},
  author={Yang, An and Li, Anfeng and Yang, Baosong and Zhang, Beichen and Hui, Binyuan and Zheng, Bo and Yu, Bowen and Gao, Chang and Huang, Chengen and Lv, Chenxu and others},
  journal={arXiv preprint arXiv:2505.09388},
  year={2025}
}

@article{comanici2025gemini,
  title={Gemini 2.5: Pushing the frontier with advanced reasoning, multimodality, long context, and next generation agentic capabilities},
  author={Comanici, Gheorghe and Bieber, Eric and Schaekermann, Mike and Pasupat, Ice and Sachdeva, Noveen and Dhillon, Inderjit and Blistein, Marcel and Ram, Ori and Zhang, Dan and Rosen, Evan and others},
  journal={arXiv preprint arXiv:2507.06261},
  year={2025}
}

@online{gpt41,
  title        = {Introducing GPT-4.1 in the API},
  author       = {OpenAI},
  year         = {2025},
  url          = {https://openai.com/index/gpt-4-1/},
}

@online{o4mini,
  title        = {Introducing o3 and o4-mini},
  author       = {OpenAI},
  year         = {2025},
  url          = {https://openai.com/index/introducing-o3-and-o4-mini/},
}

@online{claude4,
  title        = {Introducing Claude 4},
  author       = {Anthropic},
  year         = {2025},
  url          = {https://www.anthropic.com/news/claude-4},
}

@online{gemini3pro,
  title        = {Gemini 3 Pro: the frontier of vision AI},
  author       = {Google},
  year         = {2025},
  url          = {https://blog.google/innovation-and-ai/technology/developers-tools/gemini-3-pro-vision/},
}

@article{singh2025openai,
  title={Openai gpt-5 system card},
  author={Singh, Aaditya and Fry, Adam and Perelman, Adam and Tart, Adam and Ganesh, Adi and El-Kishky, Ahmed and McLaughlin, Aidan and Low, Aiden and Ostrow, AJ and Ananthram, Akhila and others},
  journal={arXiv preprint arXiv:2601.03267},
  year={2025}
}

@inproceedings{papineni2002bleu,
  title={Bleu: a method for automatic evaluation of machine translation},
  author={Papineni, Kishore and Roukos, Salim and Ward, Todd and Zhu, Wei-Jing},
  booktitle={Proceedings of the 40th annual meeting of the Association for Computational Linguistics},
  pages={311--318},
  year={2002}
}

@article{zheng2023judging,
  title={Judging llm-as-a-judge with mt-bench and chatbot arena},
  author={Zheng, Lianmin and Chiang, Wei-Lin and Sheng, Ying and Zhuang, Siyuan and Wu, Zhanghao and Zhuang, Yonghao and Lin, Zi and Li, Zhuohan and Li, Dacheng and Xing, Eric and others},
  journal={Advances in neural information processing systems},
  volume={36},
  pages={46595--46623},
  year={2023}
}

@article{nida1945linguistics,
  title={Linguistics and ethnology in translation-problems},
  author={Nida, Eugene},
  journal={Word},
  volume={1},
  number={2},
  pages={194--208},
  year={1945},
  publisher={Taylor \& Francis}
}

@article{nida1993language,
  title={Language, culture, and translating},
  author={Nida, Eugene Albert},
  journal={(No Title)},
  year={1993}
}

@article{bassnett2007culture,
  title={Culture and translation},
  author={Bassnett, Susan},
  journal={A companion to translation studies},
  volume={18},
  number={1},
  pages={13--23},
  year={2007},
  publisher={Multilingual Matters Clevedon}
}

@book{nida1964toward,
  title={Toward a science of translating: With special reference to principles and procedures involved in Bible translating},
  author={Nida, Eugene Albert},
  year={1964},
  publisher={Brill Archive}
}

@book{larson1997meaning,
  title={Meaning-based translation: A guide to cross-language equivalence},
  author={Larson, Mildred L},
  year={1997},
  publisher={Bloomsbury Publishing PLC}
}

@article{xu1980culturally,
  title={Culturally loaded words and English language teaching},
  author={Guozhang Xu},
  journal={\begin{CJK}{UTF8}{gbsn}现代外语\end{CJK}},
  volume={4},
  number={1},
  pages={925},
  year={1980}
}

@article{bader1994mona,
  title={Mona Baker, In Other Words: A Coursebook on Translation (Book Review)},
  author={Bader, Yousef},
  journal={IRAL: International Review of Applied Linguistics in Language Teaching},
  volume={32},
  number={1},
  pages={89},
  year={1994},
  publisher={J. Groos Verlag.}
}

@book{newmark1988textbook,
  title={A textbook of translation},
  author={Newmark, Peter},
  volume={66},
  year={1988},
  publisher={Prentice hall New York}
}

@article{baker1992other,
  title={In other words: A coursebook on translation},
  author={Baker, Mona},
  journal={London and New York: Routledge},
  year={1992}
}

@book{venuti1995translator,
  author    = {Venuti, Lawrence},
  title     = {The Translator's Invisibility: A History of Translation},
  year      = {1995},
  publisher = {Routledge},
  address   = {London}
}

@book{newmark1998more,
  title={More paragraphs on translation},
  author={Newmark, Peter},
  year={1998},
  publisher={Multilingual matters}
}

@article{oo2020culture,
  title={Culture through Keywords in Kenneth Wong's Translation of" Spirit Food" by Nu Nu Yi (Inwa)},
  author={Oo, Su Khine},
  journal={Journal of English Language and Linguistics},
  volume={1},
  number={2},
  pages={115--134},
  year={2020}
}

@article{Yuwei2022EnglishTO,
  title={English Translation of Culture-Loaded Words—A Corpus Based Study},
  author={PIAN Yu-wei and Chen Wei},
  journal={Journal of Literature and Art Studies},
  year={2022},
  url={https://api.semanticscholar.org/CorpusID:250109532}
}

@article{prompan2024study,
  title={A Study of Culture-Specific Items (CSIs) and Translation Strategies in The Blind Earthworm in the Labyrinth},
  author={Prompan, Jantra},
  journal={Rajapark Journal},
  volume={18},
  number={58},
  pages={212--235},
  year={2024}
}

@article{al2025cultural,
  title={Cultural Representation in Children’s Cartoon Programmes: Insights from the Nida/Newmark Typology},
  author={Al Rujaibi, Raya Nasser and Ariffin, Adlina and Jamoussi, Rafik},
  journal={International Journal of Language, Literacy and Translation},
  volume={8},
  number={2},
  year={2025}
}

@article{brown2020language,
  title={Language models are few-shot learners},
  author={Brown, Tom and Mann, Benjamin and Ryder, Nick and Subbiah, Melanie and Kaplan, Jared D and Dhariwal, Prafulla and Neelakantan, Arvind and Shyam, Pranav and Sastry, Girish and Askell, Amanda and others},
  journal={Advances in neural information processing systems},
  volume={33},
  pages={1877--1901},
  year={2020}
}

@article{touvron2023llama,
  title={Llama 2: Open foundation and fine-tuned chat models},
  author={Touvron, Hugo and Martin, Louis and Stone, Kevin and Albert, Peter and Almahairi, Amjad and Babaei, Yasmine and Bashlykov, Nikolay and Batra, Soumya and Bhargava, Prajjwal and Bhosale, Shruti and others},
  journal={arXiv preprint arXiv:2307.09288},
  year={2023}
}

@article{bai2023qwen,
  title={Qwen technical report},
  author={Bai, Jinze and Bai, Shuai and Chu, Yunfei and Cui, Zeyu and Dang, Kai and Deng, Xiaodong and Fan, Yang and Ge, Wenbin and Han, Yu and Huang, Fei and others},
  journal={arXiv preprint arXiv:2309.16609},
  year={2023}
}

@inproceedings{vilar2023prompting,
  title={Prompting palm for translation: Assessing strategies and performance},
  author={Vilar, David and Freitag, Markus and Cherry, Colin and Luo, Jiaming and Ratnakar, Viresh and Foster, George},
  booktitle={Proceedings of the 61st Annual Meeting of the Association for Computational Linguistics (Volume 1: Long Papers)},
  pages={15406--15427},
  year={2023}
}

@article{he2024exploring,
  title={Exploring human-like translation strategy with large language models},
  author={He, Zhiwei and Liang, Tian and Jiao, Wenxiang and Zhang, Zhuosheng and Yang, Yujiu and Wang, Rui and Tu, Zhaopeng and Shi, Shuming and Wang, Xing},
  journal={Transactions of the Association for Computational Linguistics},
  volume={12},
  pages={229--246},
  year={2024},
  publisher={MIT Press One Broadway, 12th Floor, Cambridge, Massachusetts 02142, USA~…}
}

@article{jiao2023chatgpt,
  title={Is ChatGPT a good translator? Yes with GPT-4 as the engine},
  author={Jiao, Wenxiang and Wang, Wenxuan and Huang, Jen-tse and Wang, Xing and Shi, Shuming and Tu, Zhaopeng},
  journal={arXiv preprint arXiv:2301.08745},
  year={2023}
}

@inproceedings{lommel2013multidimensional,
  title={Multidimensional quality metrics: a flexible system for assessing translation quality},
  author={Lommel, Arle Richard and Burchardt, Aljoscha and Uszkoreit, Hans},
  booktitle={Proceedings of Translating and the Computer 35},
  year={2013}
}

@article{abdin2025phi,
  title={Phi-4-reasoning technical report},
  author={Abdin, Marah and Agarwal, Sahaj and Awadallah, Ahmed and Balachandran, Vidhisha and Behl, Harkirat and Chen, Lingjiao and de Rosa, Gustavo and Gunasekar, Suriya and Javaheripi, Mojan and Joshi, Neel and others},
  journal={arXiv preprint arXiv:2504.21318},
  year={2025}
}

@article{chen2025reasoning,
  title={Reasoning models don't always say what they think},
  author={Chen, Yanda and Benton, Joe and Radhakrishnan, Ansh and Uesato, Jonathan and Denison, Carson and Schulman, John and Somani, Arushi and Hase, Peter and Wagner, Misha and Roger, Fabien and others},
  journal={arXiv preprint arXiv:2505.05410},
  year={2025}
}

@article{wang2025polymath,
  title={Polymath: Evaluating mathematical reasoning in multilingual contexts},
  author={Wang, Yiming and Zhang, Pei and Tang, Jialong and Wei, Haoran and Yang, Baosong and Wang, Rui and Sun, Chenshu and Sun, Feitong and Zhang, Jiran and Wu, Junxuan and others},
  journal={arXiv preprint arXiv:2504.18428},
  year={2025}
}

@book{saussy1995problem,
  title={The problem of a Chinese aesthetic},
  author={Saussy, Haun},
  year={1995},
  publisher={Stanford University Press}
}

@book{plaks2015archetype,
  title={Archetype and Allegory in the" Dream of the Red Chamber"},
  author={Plaks, Andrew H},
  year={2015},
  publisher={Princeton University Press}
}

@article{Edwards2005FictionsOE,
  title={Fictions of Enlightenment: Journey to the West, Tower of Myriad Mirrors and Dream of the Red Chamber, and: Androgyny in Late Ming and Early Qing Literature (review)},
  author={Louise Edwards},
  journal={China Review International},
  year={2005},
  volume={12},
  pages={154 - 156},
  url={https://api.semanticscholar.org/CorpusID:144146461}
}

@article{Xia2011OnET,
  title={On English Translation of Culture-Specific Items in the Ancient Chinese Official System:A Descriptive and Comparative Study on Hawkes’ and Yangs’ English Translated Cases of Hong Lou Meng},
  author={Qing Xia},
  journal={Cross-cultural Communication},
  year={2011},
  volume={6},
  pages={58-73},
  url={https://api.semanticscholar.org/CorpusID:162099077}
}

@book{moratto2022dream,
  title={Dream of the Red Chamber: Literary and translation perspectives},
  author={Moratto, Riccardo and Liu, Kanglong and Chao, Di-kai},
  year={2022},
  publisher={Taylor \& Francis}
}

@article{ WJYY201901011,
author = { Dan Song },
title = {\begin{CJK}{UTF8}{gbsn}《红楼梦》在日本的翻译与影响研究\end{CJK}},
journal = {\begin{CJK}{UTF8}{gbsn}外语教学与研究\end{CJK}},
volume = {51},
number = {01},
pages = {121-132+161},
year = {2019},
issn = {1000-0429},
}

@article{wan1998,
  title={\begin{CJK}{UTF8}{min}伊藤漱平訳,『紅楼夢』, 平凡社ラーブラリー, 全十二冊, 平成九年十一月完結\end{CJK}},
  author={Hiroaki Maruyama},
  journal={\begin{CJK}{UTF8}{min}二松学舎大学人文論叢\end{CJK}},
  volume={61},
  pages={102},
  year={1998},
  publisher={\begin{CJK}{UTF8}{min}二松學舎大学\end{CJK}}
}

@inproceedings{Bassnett1990TranslationHA,
  title={Translation, History and Culture},
  author={Susan Bassnett and Andr{\'e} Lef{\`e}vere},
  year={1990},
  url={https://api.semanticscholar.org/CorpusID:170422906}
}

@inproceedings{Toury1995DescriptiveTS,
  title={Descriptive translation studies and beyond},
  author={Gideon Toury},
  year={1995},
  url={https://api.semanticscholar.org/CorpusID:261678043}
}

@article{vieira2021understanding,
  title={Understanding the societal impacts of machine translation: a critical review of the literature on medical and legal use cases},
  author={Vieira, Lucas Nunes and O’Hagan, Minako and O’Sullivan, Carol},
  journal={Information, Communication \& Society},
  volume={24},
  number={11},
  pages={1515--1532},
  year={2021},
  publisher={Taylor \& Francis}
}

@article{moghe2025machine,
  title={Machine translation meta evaluation through translation accuracy challenge sets},
  author={Moghe, Nikita and Fazla, Arnisa and Amrhein, Chantal and Kocmi, Tom and Steedman, Mark and Birch, Alexandra and Sennrich, Rico and Guillou, Liane},
  journal={Computational Linguistics},
  volume={51},
  number={1},
  pages={73--137},
  year={2025}
}

@inproceedings{shen2025liaozhai,
  title={Liaozhai through the Looking-Glass: On Paratextual Explicitation of Culture-Bound Terms in Machine Translation},
  author={Shen, Sherrie and Wang, Weixuan and Birch, Alexandra},
  booktitle={Proceedings of the 2025 Conference on Empirical Methods in Natural Language Processing},
  pages={34388--34404},
  year={2025}
}

@inproceedings{zhang2025good,
  title={How good are LLMs for literary translation, really? Literary translation evaluation with humans and LLMs},
  author={Zhang, Ran and Zhao, Wei and Eger, Steffen},
  booktitle={Proceedings of the 2025 Conference of the Nations of the Americas Chapter of the Association for Computational Linguistics: Human Language Technologies (Volume 1: Long Papers)},
  pages={10961--10988},
  year={2025}
}

@article{wei2022chain,
  title={Chain-of-thought prompting elicits reasoning in large language models},
  author={Wei, Jason and Wang, Xuezhi and Schuurmans, Dale and Bosma, Maarten and Xia, Fei and Chi, Ed and Le, Quoc V and Zhou, Denny and others},
  journal={Advances in neural information processing systems},
  volume={35},
  pages={24824--24837},
  year={2022}
}

@inproceedings{aixela1996culture,
  title={Culture-specific items in translation},
  author={Aixel{\'a}, Javier Franco},
  booktitle={Translation, power, subversion},
  pages={52--78},
  year={1996},
  organization={Multilingual Matters}
}

@book{wu2019yanshi,
  author    = {Wu, Jun},
  title     = {\begin{CJK}{UTF8}{gbsn}阐释的演化：伊藤漱平《红楼梦》日译研究\end{CJK}},
  year      = {2019},
  address   = {\begin{CJK}{UTF8}{gbsn}北京\end{CJK}},
  publisher = {\begin{CJK}{UTF8}{gbsn}知识产权出版社\end{CJK}},
  note      = {167--168}
}

@article{sun2006,
  title={\begin{CJK}{UTF8}{gbsn}日本《红楼梦》研究略史\end{CJK}},
  author={Sun, Yuming},
  journal={\begin{CJK}{UTF8}{gbsn}红楼梦学刊\end{CJK}},
  number={5},
  pages={224--240},
  year={2006}
}

@book{Hu1993,
  author    = {Hu, Wenbin},
  title     = {\begin{CJK}{UTF8}{gbsn}《红楼梦》在国外\end{CJK}},
  publisher = {\begin{CJK}{UTF8}{gbsn}中华书局\end{CJK}},
  address   = {\begin{CJK}{UTF8}{gbsn}北京\end{CJK}},
  year      = {1993},
  month     = {11},
  pages     = {1-25}
}

@inproceedings{kocmi2024findings,
  title={Findings of the WMT24 general machine translation shared task: The LLM era is here but MT is not solved yet},
  author={Kocmi, Tom and Avramidis, Eleftherios and Bawden, Rachel and Bojar, Ond{\v{r}}ej and Dvorkovich, Anton and Federmann, Christian and Fishel, Mark and Freitag, Markus and Gowda, Thamme and Grundkiewicz, Roman and others},
  booktitle={Proceedings of the Ninth Conference on Machine Translation},
  pages={1--46},
  year={2024}
}

@article{pang2025salute,
  title={Salute the classic: Revisiting challenges of machine translation in the age of large language models},
  author={Pang, Jianhui and Ye, Fanghua and Wong, Derek Fai and Yu, Dian and Shi, Shuming and Tu, Zhaopeng and Wang, Longyue},
  journal={Transactions of the Association for Computational Linguistics},
  volume={13},
  pages={73--95},
  year={2025},
  publisher={MIT Press 255 Main Street, 9th Floor, Cambridge, Massachusetts 02142, USA~…}
}

@article{ataman2025machine,
  title={Machine translation in the era of large language models: a survey of historical and emerging problems},
  author={Ataman, Duygu and Birch, Alexandra and Habash, Nizar and Federico, Marcello and Koehn, Philipp and Cho, Kyunghyun},
  journal={Information},
  volume={16},
  number={9},
  pages={723},
  year={2025},
  publisher={MDPI}
}

@article{riley2023frmt,
  title={Frmt: A benchmark for few-shot region-aware machine translation},
  author={Riley, Parker and Dozat, Timothy and Botha, Jan A and Garcia, Xavier and Garrette, Dan and Riesa, Jason and Firat, Orhan and Constant, Noah},
  journal={Transactions of the Association for Computational Linguistics},
  volume={11},
  pages={671--685},
  year={2023},
  publisher={MIT Press One Broadway, 12th Floor, Cambridge, Massachusetts 02142, USA~…}
}

@inproceedings{hershcovich2022challenges,
  title={Challenges and strategies in cross-cultural NLP},
  author={Hershcovich, Daniel and Frank, Stella and Lent, Heather and De Lhoneux, Miryam and Abdou, Mostafa and Brandl, Stephanie and Bugliarello, Emanuele and Piqueras, Laura Cabello and Chalkidis, Ilias and Cui, Ruixiang and others},
  booktitle={Proceedings of the 60th Annual Meeting of the Association for Computational Linguistics (Volume 1: Long Papers)},
  pages={6997--7013},
  year={2022}
}

@inproceedings{yao2024benchmarking,
  title={Benchmarking machine translation with cultural awareness},
  author={Yao, Binwei and Jiang, Ming and Bobinac, Tara and Yang, Diyi and Hu, Junjie},
  booktitle={Findings of the Association for Computational Linguistics: EMNLP 2024},
  pages={13078--13096},
  year={2024}
}

@article{cao2024cultural,
  title={Cultural adaptation of recipes},
  author={Cao, Yong and Kementchedjhieva, Yova and Cui, Ruixiang and Karamolegkou, Antonia and Zhou, Li and Dare, Megan and Donatelli, Lucia and Hershcovich, Daniel},
  journal={Transactions of the Association for Computational Linguistics},
  volume={12},
  pages={80--99},
  year={2024},
  publisher={MIT Press One Broadway, 12th Floor, Cambridge, Massachusetts 02142, USA~…}
}

@inproceedings{zhang2024cultural,
  title={Cultural adaptation of menus: A fine-grained approach},
  author={Zhang, Zhonghe and He, Xiaoyu and Iyer, Vivek and Birch, Alexandra},
  booktitle={Proceedings of the Ninth Conference on Machine Translation},
  pages={1258--1271},
  year={2024}
}

@inproceedings{pandey2025culturally,
  title={CULTURALLY YOURS: A reading assistant for cross-cultural content},
  author={Pandey, Saurabh Kumar and Budhiraja, Harshit and Saha, Sougata and Choudhury, Monojit},
  booktitle={Proceedings of the 31st International Conference on Computational Linguistics: System Demonstrations},
  pages={208--216},
  year={2025}
}

\appendix

\section{More Related Work}
\label{appe:related-work}

Regarding related work on culturally loaded translation, \textcolor{deepred}{§\ref{sec:foundations}} has already discussed foundational theoretical studies. Here, we focus only on MT research that potentially relates to our work.

MT research has long recognized the challenges of culture-oriented translation \citep{hershcovich2022challenges}.
In response, several studies have proposed benchmarks to quantify cultural translation \citep{riley2023frmt,cao2024cultural,yao2024benchmarking,zhang2024cultural}, but these often focus on specific domains or concrete contexts (e.g., food) and rely primarily on automatic metrics.
Other work has developed systems for identifying culturally loaded content \citep{pandey2025culturally}.
\citet{shen2025liaozhai} shares a similar perspective with ours, namely rethinking the challenges of culturally loaded translation in MT, but focuses specifically on the aspect of paratexts.
In contrast, our study adopts a more systematic perspective, offering a comprehensive analysis of culturally loaded translation challenges from the task itself to its evaluation.

\section{Dataset}
\label{appe:data}

\subsection{More Examples}
\label{appe:examples}

\textcolor{deepred}{Tab.\ref{tab:cases-ecology}} - \ref{tab:cases-society} presents dataset examples from five cultural categories.

\begin{table*}[htbp]
\caption{Dataset examples of the {\bf Ecology} category.}
\vspace{-0.1in}
\centering
\footnotesize
\renewcommand\arraystretch{1.1}
\setlength{\tabcolsep}{2.5mm}{
\resizebox{1.95\columnwidth}{!}{
\begin{tabular}{p{15cm}}
\toprule

{\bf Source:}
\begin{CJK}{UTF8}{gbsn}
每日早起，拿上等燕窝一两，冰糖五钱，用银铫子熬出粥来。
\end{CJK}
\\
{\bf Target:}
\begin{CJK}{UTF8}{gbsn}
毎朝、上等の燕窩（海つばめの巣。燕巣）一両分と氷砂糖五銭分とを、銀の銚子で煮いてお粥にこしらえなさい。
\end{CJK}
\\[10pt]

{\bf Source:}
\begin{CJK}{UTF8}{gbsn}
我不吃六安茶
\end{CJK}
\\
{\bf Target:}
\begin{CJK}{UTF8}{gbsn}
「わしは六安茶（安徽省霍山県---昔、六安郡に属した---産の銘茶）は飲まないのだよ」
\end{CJK}
\\[10pt]

{\bf Source:}
\begin{CJK}{UTF8}{gbsn}
此刻忽见宝玉笑问道：“宝姐姐，我瞧瞧你的红麝串子？
\end{CJK}
\\
{\bf Target:}
\begin{CJK}{UTF8}{gbsn}
宝玉はにこにこしながら、「宝釵お姉さま、あなたの分の赤い香りの香珠、わたしに拝ませてくださいよ」と言い出すのでした。
\end{CJK}
\\[20pt]

{\bf Source:}
\begin{CJK}{UTF8}{gbsn}
只见晴雯如得了世露一般，一气都灌下去了。
\end{CJK}
\\
{\bf Target:}
\begin{CJK}{UTF8}{gbsn}
すると晴雯はさながら甘露でも得たかのように、一気にこれを咽喉に流しこむのでした。
\end{CJK}
\\[10pt]

{\bf Source:}
\begin{CJK}{UTF8}{gbsn}
刘姥姥道：“这个菜里有毒，俺们那些都成了砒霜了。那怕毒死了，也要吃尽了。”
\end{CJK}
\\
{\bf Target:}
\begin{CJK}{UTF8}{gbsn}
劉婆さんは答えて、「このご馳走に毒がはいっておるちゅうことになりますると、てまえどもの料理なんぞは、砒霜（砒素の化合物、毒物）の塊ってえことになっちまいますわい。たとい毒で死のうとも、ご馳走さえ平らげられたら本望でございます」
\end{CJK}
\\[10pt]

\bottomrule
\end{tabular}}}
\label{tab:cases-ecology}
\vspace{-0.in}
\end{table*}


\begin{table*}[htbp]
\caption{Dataset examples of the {\bf Religion} category.}
\vspace{-0.1in}
\centering
\footnotesize
\renewcommand\arraystretch{1.1}
\setlength{\tabcolsep}{2.5mm}{
\resizebox{1.95\columnwidth}{!}{
\begin{tabular}{p{15cm}}
\toprule

{\bf Source:}
\begin{CJK}{UTF8}{gbsn}
贾珍便命贾琼，责琛、贾璘、贾蔷四个人去陪客，一面吩咐去请钦天监阴阳司来择日。
\end{CJK}
\\
{\bf Target:}
\begin{CJK}{UTF8}{gbsn}
に言いつけて客の応対に当たらせることとし、一方では使いをやって欽天監（天文暦法をつかさどる役所）の陰陽司の係（陰陽生、後出。易占をつかさどる）に日柄を見たてさせます。
\end{CJK}
\\[15pt]

{\bf Source:}
\begin{CJK}{UTF8}{gbsn}
风姐听了这话，便发了兴头，说道：“你是素日知道我的，从来不信什么是阴司地狱报应的。
\end{CJK}
\\
{\bf Target:}
\begin{CJK}{UTF8}{gbsn}
無鳳はこのことばを聞いてつい釣りこまれ、「あなたはかねがねこのわたしという人間をご存じのはず、閻魔の庁だの地獄の応報だのはついぞじたこともないわたしですもの」
\end{CJK}
\\[15pt]

{\bf Source:}
\begin{CJK}{UTF8}{gbsn}
“我昨日叫赖升媳妇出去叫人给宝玉算算命。这先生算得好灵，说要娶了金命的人帮扶他，必要冲冲喜才好。不然只怕保不住。”
\end{CJK}
\\
{\bf Target:}
\begin{CJK}{UTF8}{gbsn}
わしは昨日、頼昇の家内を街へつかわして宝玉のことを占わせてみたのだが、その占い者の立てた卦というのがぴったりなのだよ---金の性の者を嫁にとって連れ添わせなさい。どうしてもここでひとつ縁起直しせぬことにはいかん。でないと恐らくは生命も保つまい、とこういったそうだ。
\end{CJK}
\\[25pt]


{\bf Source:}
\begin{CJK}{UTF8}{gbsn}
你舅舅今日斋戒去了
\end{CJK}
\\
{\bf Target:}
\begin{CJK}{UTF8}{gbsn}
伯父さんは今日はお役で斎戒にお出かけになってお留守なの。
\end{CJK}
\\

\bottomrule
\end{tabular}}}
\label{tab:cases-religion}
\vspace{-0.in}
\end{table*}


\begin{table*}[htbp]
\caption{Dataset examples of the {\bf Material} category.}
\vspace{-0.1in}
\centering
\footnotesize
\renewcommand\arraystretch{1.1}
\setlength{\tabcolsep}{2.5mm}{
\resizebox{1.95\columnwidth}{!}{
\begin{tabular}{p{15cm}}
\toprule

{\bf Source:}
\begin{CJK}{UTF8}{gbsn}
王子腾那边，仍是一套衣服，一双鞋袜，一百寿桃，一百束上用银丝挂面；
\end{CJK}
\\
{\bf Target:}
\begin{CJK}{UTF8}{gbsn}
王子騰のもとからは、例によって衣服一襲・靴下一足・寿桃（誕生祝いに送る小麦粉の桃）百個・宮中用素麵百束が届きます
\end{CJK}
\\[15pt]

{\bf Source:}
\begin{CJK}{UTF8}{gbsn}
这红玉也不梳洗，向镜中胡乱挽了一挽头发，洗了洗手，腰内束了一条汗巾子便来打扫房屋。
\end{CJK}
\\
{\bf Target:}
\begin{CJK}{UTF8}{gbsn}
紅玉とておちおち身づくろいなどしてはおられず、そうそうに鏡に向かって髪をわがね、手を洗い、腰帯をしめるなり、部屋の掃除にやってきました。
\end{CJK}
\\[15pt]

{\bf Source:}
\begin{CJK}{UTF8}{gbsn}
宝琴披着凫靥裘站在那里笑
\end{CJK}
\\
{\bf Target:}
\begin{CJK}{UTF8}{gbsn}
鳧の毛の装をはおった宝琴がそこに立って笑っています。
\end{CJK}
\\[10pt]

{\bf Source:}
\begin{CJK}{UTF8}{gbsn}
另换了三四个衣帽周全十七八岁的小厮上来，复抬起轿子，众婆五步下围随，至一垂花门前落下。
\end{CJK}
\\
{\bf Target:}
\begin{CJK}{UTF8}{gbsn}
するとこんどは別に三、四人、お仕着せ姿をした十七、八の若党が交替にきて、また輪をかきあげ、老女たちがその囲りをとりまくようにして徒歩でついてゆき、垂花門（正門をはいった中門、二の門。垂花のかざりがあるのでいう）まできて幅をおろしました。
\end{CJK}
\\[25pt]

{\bf Source:}
\begin{CJK}{UTF8}{gbsn}
散押岁钱荷包金银锞；摆上合欢宴来，男东女西归坐，献屠苏酒、合欢汤、吉祥果、如意糕毕。
\end{CJK}
\\
{\bf Target:}
\begin{CJK}{UTF8}{gbsn}
そこで押歳銭（大晦日に長上から子供に取らせるお年玉の金。穴あき銅銭百枚を赤紙で通してある）・中着・金銀の小粒などを分けて取らせます。また合歓宴の宴席を設けて、男は東の、女は西の席につき、屠蘇酒・合歓湯（スープ）・吉祥果（果物）・如意糕（蒸し菓子）を献じ終えました。
\end{CJK}
\\

\bottomrule
\end{tabular}}}
\label{tab:cases-material}
\vspace{-0.in}
\end{table*}


\begin{table*}[htbp]
\caption{Dataset examples of the {\bf Linguistics} category.}
\vspace{-0.1in}
\centering
\footnotesize
\renewcommand\arraystretch{1.1}
\setlength{\tabcolsep}{2.5mm}{
\resizebox{1.95\columnwidth}{!}{
\begin{tabular}{p{15cm}}
\toprule

{\bf Source:}
\begin{CJK}{UTF8}{gbsn}
谋事在人，成事在天
\end{CJK}
\\
{\bf Target:}
\begin{CJK}{UTF8}{gbsn}
人間さまがお膳立て、天道さまがお取りあげ
\end{CJK}
\\[10pt]

{\bf Source:}
\begin{CJK}{UTF8}{gbsn}
那宝玉是个丈八的灯台，照见人家，照不见自家的。
\end{CJK}
\\
{\bf Target:}
\begin{CJK}{UTF8}{gbsn}
どだい肝腎の宝玉さまが、『文八（一文八尺）のお灯明台：他人は照らせても、わが身は照らせぬ（「灯台もと暗し」）』。
\end{CJK}
\\[15pt]

{\bf Source:}
\begin{CJK}{UTF8}{gbsn}
巧媳妇做不出没米的粥来
\end{CJK}
\\
{\bf Target:}
\begin{CJK}{UTF8}{gbsn}
遣繰り上手の嫁さんでも米なしでは粥はできぬ
\end{CJK}
\\[10pt]

{\bf Source:}
\begin{CJK}{UTF8}{gbsn}
偏偏凤姐想出一条偷梁换柱之计
\end{CJK}
\\
{\bf Target:}
\begin{CJK}{UTF8}{gbsn}
あいにく鳳ちゃんが替え玉の計略を考え出してくれた
\end{CJK}
\\[10pt]

{\bf Source:}
\begin{CJK}{UTF8}{gbsn}
于是尤氏一行人悄悄的来至窗下，只听里面称三赞四，耍笑之音虽多；又兼着恨五骂六，忿怨之声亦不少。
\end{CJK}
\\
{\bf Target:}
\begin{CJK}{UTF8}{gbsn}
かくて尤氏ら一行、足音を忍ばせて窓の下までやってきましたところ、なかではほめたりたたえたりで笑いさんざめく声がしきりにする一方、またわめいたり恨んだりの怒りと憎しみの声も少なくないふう・・・・・
\end{CJK}
\\

\bottomrule
\end{tabular}}}
\label{tab:cases-linguistics}
\vspace{-0.in}
\end{table*}


\begin{table*}[htbp]
\caption{Dataset examples of the {\bf Society} category.}
\vspace{-0.1in}
\centering
\footnotesize
\renewcommand\arraystretch{1.1}
\setlength{\tabcolsep}{2.5mm}{
\resizebox{1.95\columnwidth}{!}{
\begin{tabular}{p{15cm}}
\toprule

{\bf Source:}
\begin{CJK}{UTF8}{gbsn}
我们合家大小登门去磕头。
\end{CJK}
\\
{\bf Target:}
\begin{CJK}{UTF8}{gbsn}
わたくしども家中揃ってお宅に伺い叩頭させていただきます
\end{CJK}
\\[10pt]

{\bf Source:}
\begin{CJK}{UTF8}{gbsn}
他是我们这里有名的一个泼皮破落户儿，南省俗谓作辣子。
\end{CJK}
\\
{\bf Target:}
\begin{CJK}{UTF8}{gbsn}
これはうちでは聞こえたお転婆の破落戸、江南の方なら俗に『辣子』というやつよ
\end{CJK}
\\[15pt]

{\bf Source:}
\begin{CJK}{UTF8}{gbsn}
他又成了香饽饽了，都抢不到手。
\end{CJK}
\\
{\bf Target:}
\begin{CJK}{UTF8}{gbsn}
あのひとはほかほか饅頭（人気者の意）なってしまい、奪い合いでめったに手にははいらないね
\end{CJK}
\\[10pt]

{\bf Source:}
\begin{CJK}{UTF8}{gbsn}
平儿便福下去，宝玉作揖不迭。
\end{CJK}
\\
{\bf Target:}
\begin{CJK}{UTF8}{gbsn}
平児がそこで「万福（女子の敬礼のときのことば）」といってお辞儀をしますと、宝玉は遅れじ揖礼（手を挟いて上下する男子の敬礼法）を返します。
\end{CJK}
\\[25pt]

{\bf Source:}
\begin{CJK}{UTF8}{gbsn}
“孽障！你生气，要打骂人容易，何苦摔那命根子！”
\end{CJK}
\\
{\bf Target:}
\begin{CJK}{UTF8}{gbsn}
「この罰あたりめが！おまえ、かんしゃくを起こしたら、人をぶつたりどなりつけたりするのだってたやすいのに、選りに選ってその命の綱も同然の品を投げつけてなんとする？」
\end{CJK}
\\

\bottomrule
\end{tabular}}}
\label{tab:cases-society}
\vspace{-0.in}
\end{table*}

\subsection{Original PDF Format}
\label{appe:pdf}

\begin{figure*}[htbp]
  \centering
  \includegraphics[width=1.8\columnwidth]{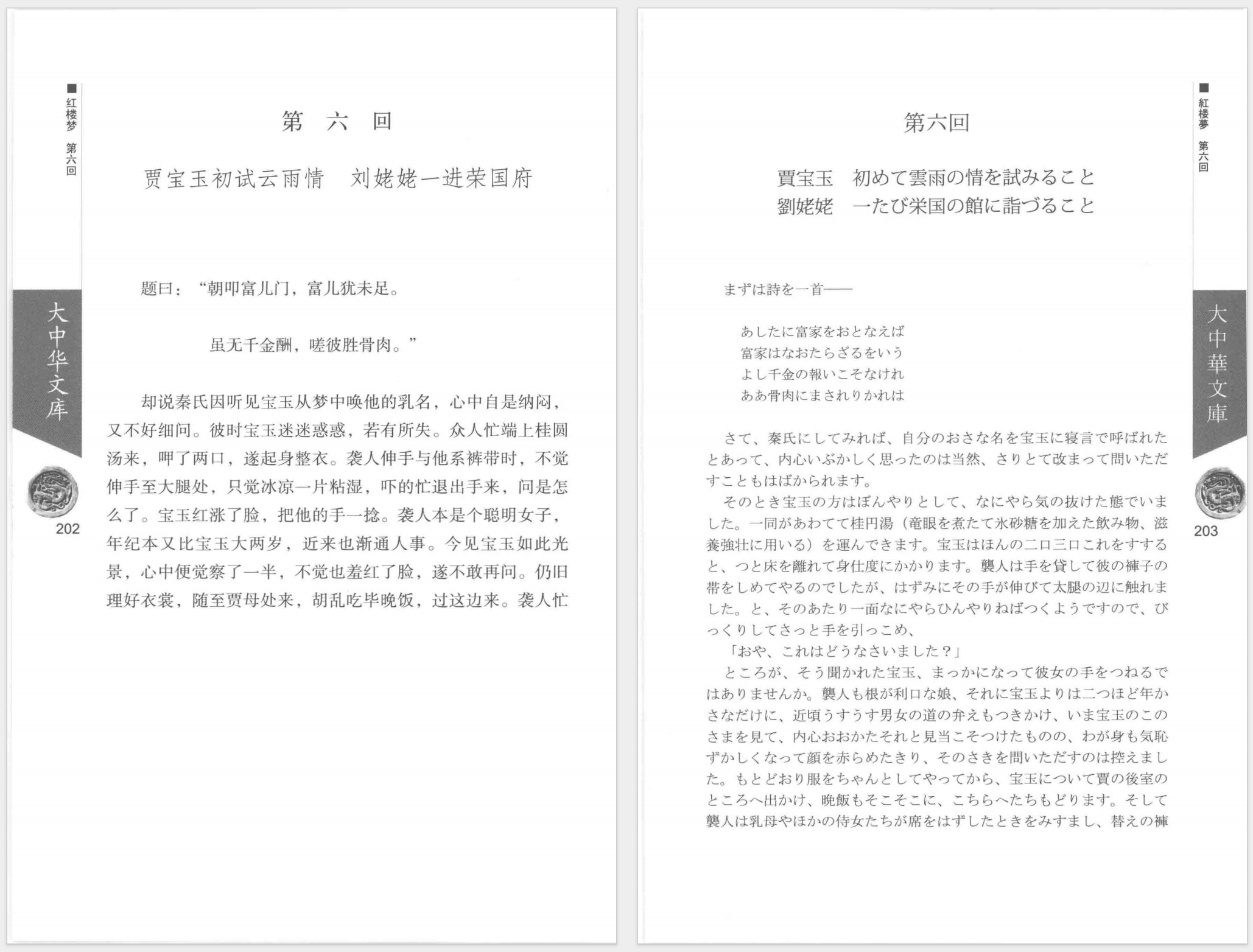} 
  \vspace{-0.in}
  \caption{Original PDF format of our corpus with direct bilingual comparisons (Case 1).}
  \label{fig:pdf1}
\end{figure*}

\begin{figure*}[htbp]
  \centering
  \includegraphics[width=1.8\columnwidth]{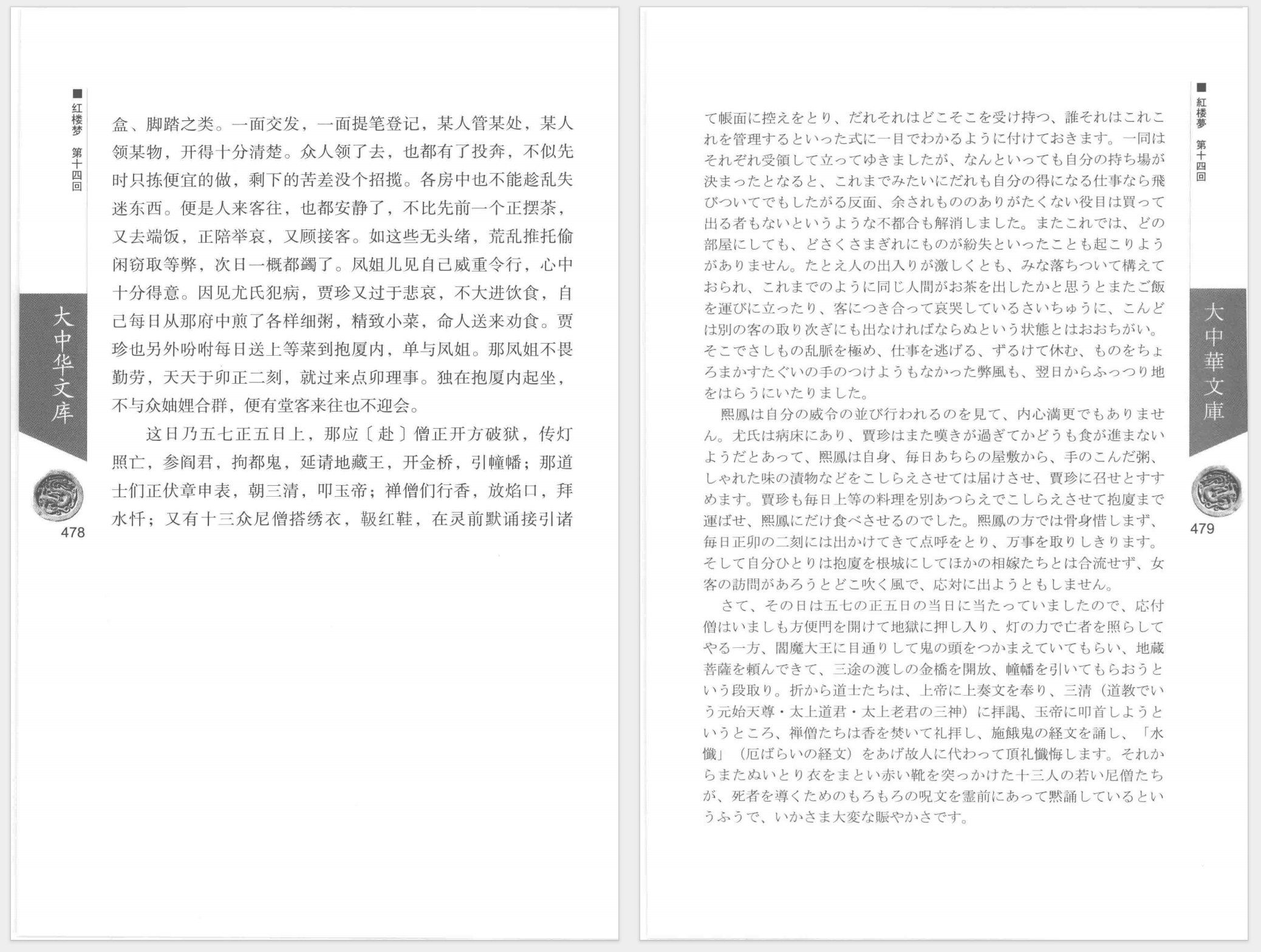} 
  \vspace{-0.in}
  \caption{Original PDF format of our corpus with direct bilingual comparisons (Case 2).}
  \label{fig:pdf2}
\end{figure*}

\textcolor{deepred}{Fig.\ref{fig:pdf1}} and \ref{fig:pdf2} present two cases of the original PDF format of our corpus with direct bilingual comparisons.

\section{Model Usage}

\subsection{Inference Implementation}
\label{appe:implementation}

\begin{table}[htbp]
\caption{Inference hyperparameters used in our experiments.}
\vspace{-0.1in}
\centering
\footnotesize
\renewcommand\arraystretch{1.1}
\setlength{\tabcolsep}{1.8mm}{
\resizebox{1\columnwidth}{!}{
\begin{tabular}{l|ccc}

\toprule
Model & Top-$k$ & Top-$p$ & Temperature $T$ \\
\midrule

Deepseek-v3 & 20 & 0.95 & 0.6 \\
Deepseek-r1 & 20 & 0.95 & 0.6 \\
Qwen3-235B-A22B-Non-Thinking & 20 & 0.8 & 0.7 \\
Qwen3-235B-A22B-Thinking & 20 & 0.95 & 0.6 \\

\bottomrule

\end{tabular}}}
\label{tab:hyper}
\vspace{-0.in}
\end{table}

All model implementations follow default settings. For closed-source models, we do not have access to hyperparameters. For open-source models, specific sampling hyperparameters are listed in \textcolor{deepred}{Tab.\ref{tab:hyper}}.

Given the randomness of sampling, theoretically, we should generate multiple responses and report the average. However, due to the high cost of human evaluation, both in time and expense, it is impractical to score every sampled output.
To address this, we randomly selected 20 samples, had each model generate eight responses, and scored them separately by our human evaluators.
We observed that score variance across samples was minimal, unlike in deterministic reasoning tasks, where outputs can vary significantly \citep{abdin2025phi,chen2025reasoning,wang2025polymath}.
Under this premise, we use only the first sampled response for evaluation.

\subsection{LLM-as-a-Judge Prompt}
\label{appe:llm-judge}

\textcolor{deepred}{Tab.\ref{tab:llm-judge}} presents the prompt used for our llm-as-a-judge evaluations.

\begin{table*}[htbp]
\caption{Prompt used for LLM-as-a-judge evaluation.}
\vspace{-0.1in}
\centering
\footnotesize
\renewcommand\arraystretch{1.1}
\setlength{\tabcolsep}{2.5mm}{
\resizebox{1.95\columnwidth}{!}{
\begin{tabular}{p{15cm}}
\toprule

\#\#\# Instruction:\\
You are an expert in translation evaluation with native-level proficiency in both source and target languages and deep knowledge of both cultures. You will evaluate a translation based on the following four dimensions. After considering each dimension, provide an **Overall Score** from 0 to 5 that holistically reflects the translation's quality.\\[25pt]

---\\[10pt]

\#\#\# Evaluation Dimensions:\\[10pt]

**1. Content Accuracy**  \\
Measures whether the translation faithfully and accurately conveys the meaning of the source text.  \\
- High accuracy means no noticeable deviation from the original meaning.  \\
- Low accuracy indicates mistranslations, distortions, or loss of core content.\\[10pt]

**2. Language Fluency**  \\
Measures whether the translation conforms to the linguistic norms of the target language, including grammar, vocabulary, spelling, punctuation, and naturalness of expression.  \\
- Fluent translations read smoothly and naturally.  \\
- Low fluency involves awkward wording, grammatical errors, or obvious translationese.\\[10pt]

**3. Cultural Appropriateness**  \\
Evaluates whether appropriate translation strategies are used when handling cultural elements in the source text, ensuring that cultural connotations are properly conveyed and understandable to target readers.  \\
- High cultural appropriateness preserves cultural nuances without causing confusion.  \\
- Low appropriateness results in cultural loss, distortion, or misunderstanding.\\[10pt]

**4. Native Readability**  \\
Evaluates whether the translation sounds natural and acceptable to native speakers of the target language and can be easily understood without barriers.  \\
- High readability means the text is clear and fully understandable at first reading.  \\
- Low readability requires inference, repeated reading, or leaves only a general impression.\\[10pt]

---\\[10pt]

\#\#\# Overall Scoring Criteria:\\
- **5 - Excellent**: Highly accurate, fluent, culturally appropriate, and reads naturally for native speakers. No noticeable issues.\\
- **4 - Good**: Generally accurate and fluent with minor deviations; cultural meaning largely preserved; occasional awkwardness but easily understood.\\
- **3 - Adequate**: Main content is understandable, but noticeable issues exist in accuracy, fluency, or cultural handling; may require some effort to read.\\
- **2 - Poor**: Significant deviations or errors; cultural meaning poorly conveyed; awkward or hard to follow; only general ideas graspable.\\
- **1 - Very Poor**: Severe distortions or omissions; cultural content missing; highly unnatural or unintelligible.\\
- **0 - Completely Wrong**: Translation is unrelated to the source or empty.\\[10pt]

---\\[10pt]

\#\#\# Source Text:\\
\{source\_text\}\\[10pt]

\#\#\# Translation:\\
\{target\_text\}\\[10pt]

\#\#\# Evaluation:\\
Provide a **brief justification** (one or two sentences) and then the **Overall Score** in the following format:\\[10pt]

Justification: [your reasoning]  \\
Overall Score: [0-5]\\

\bottomrule
\end{tabular}}}
\label{tab:llm-judge}
\vspace{-0.in}
\end{table*}

\section{Human Guideline}
\label{appe:main}

\textcolor{deepred}{Tab.\ref{tab:evaluation_criteria}} presents the explanations and scoring guidelines of four specific evaluation dimensions and one overall dimension used in our experiments.

\textcolor{red}{\textbf{Note:}} During human evaluation, all evaluators were presented with guidelines in their respective native languages.
The English version is provided in this section for ease of presentation.

\begin{table*}[t]
\centering
\small
\caption{Human evaluation criteria used in our experiments.}
\label{tab:evaluation_criteria}

\renewcommand\arraystretch{1.15}

\begin{tabularx}{\textwidth}{
p{1.5cm}
p{4.5cm}
>{\raggedright\arraybackslash}X
}

\toprule
\textbf{Dimension} & \textbf{Definition} & \textbf{Scoring Criteria} \\
\midrule

\textbf{Content Accuracy}
&
Measures whether the translation faithfully and accurately conveys the meaning of the source text.
&
\makecell[tl]{
\textbf{5}: Highly accurate with no noticeable deviation.\\
\textbf{4}: Generally accurate with minor distortion.\\
\textbf{3}: Some deviations exist, but the main content is understandable.\\
\textbf{2}: Significant deviation; core content difficult to understand.\\
\textbf{1}: Severely deviates from the source text; core meaning completely lost.
}
\\

\midrule

\textbf{Language Fluency}
&
Measures whether the translation conforms to the linguistic norms of the target language, including grammar, vocabulary, spelling, punctuation, and naturalness of expression.
&
\makecell[tl]{
\textbf{5}: Smooth and natural; linguistically well-formed.\\
\textbf{4}: Generally fluent with slight unnaturalness.\\
\textbf{3}: Basically understandable but with noticeable issues.\\
\textbf{2}: Awkward wording; difficult to understand or obvious translationese.\\
\textbf{1}: Frequent errors; difficult to understand.
}
\\

\midrule

\textbf{Cultural Appropriateness}
&
Evaluates whether appropriate translation strategies are used when handling cultural elements in the source text, ensuring that cultural connotations are properly conveyed and understandable to target readers.
&
\makecell[tl]{
\textbf{5}: Cultural meanings are accurately conveyed and easily understood.\\
\textbf{4}: Generally fluent and appropriate with minor deviations.\\
\textbf{3}: Cultural information largely preserved but with some mistranslation.\\
\textbf{2}: Cultural meaning poorly conveyed, causing misunderstanding.\\
\textbf{1}: Cultural content missing or seriously distorted.
}
\\

\midrule

\textbf{Native Readability}
&
Evaluates whether the translation sounds natural and acceptable to native speakers of the target language and can be easily understood without barriers.
&
\makecell[tl]{
\textbf{5}: Meaning is clear and fully understandable.\\
\textbf{4}: Overall natural with minor awkwardness.\\
\textbf{3}: Comprehension requires inference or repeated reading.\\
\textbf{2}: Most content obscure; only general idea graspable.\\
\textbf{1}: Highly unnatural; difficult for native readers.
}
\\

\midrule

\textbf{Overall Score}
&
Provides a comprehensive evaluation of the translation’s overall quality by considering all aspects above.
&
\makecell[tl]{
\textbf{5}: Excellent overall performance across all aspects.\\
\textbf{4}: Generally good with minor shortcomings.\\
\textbf{3}: Moderate quality with noticeable issues.\\
\textbf{2}: Poor overall quality with significant problems.\\
\textbf{1}: Extremely poor overall performance.
}
\\

\bottomrule

\end{tabularx}
\label{tab:dimension-detail}
\end{table*}

\begin{table*}[t]
\caption{Error types and corresponding descriptions.}
\vspace{-0.1in}
\centering
\footnotesize
\renewcommand\arraystretch{1.1}
\setlength{\tabcolsep}{1.8mm}{
\resizebox{2\columnwidth}{!}{
\begin{tabular}{lp{10cm}}

\toprule
\textbf{Category} & \textbf{Description} \\
\midrule
{\bf A. No obvious error} & The translation accurately conveys the source content without any issues. \\
\addlinespace
{\bf B. Mistranslation} & The content is incorrectly understood, resulting in a translation that deviates from the original meaning. \\
\addlinespace
{\bf C. Overtranslation} & Information not present in the source text is added to the translation. \\
\addlinespace
{\bf D. Undertranslation} & Portions of the source text are left untranslated and directly copied into the target text. \\
\addlinespace
{\bf E. Omission} & Source information is omitted, resulting in an incomplete translation. \\
\bottomrule

\end{tabular}}}
\label{tab:error-type}
\vspace{-0.in}
\end{table*}

\begin{figure*}[t]
\centering
\includegraphics[width=2\columnwidth]{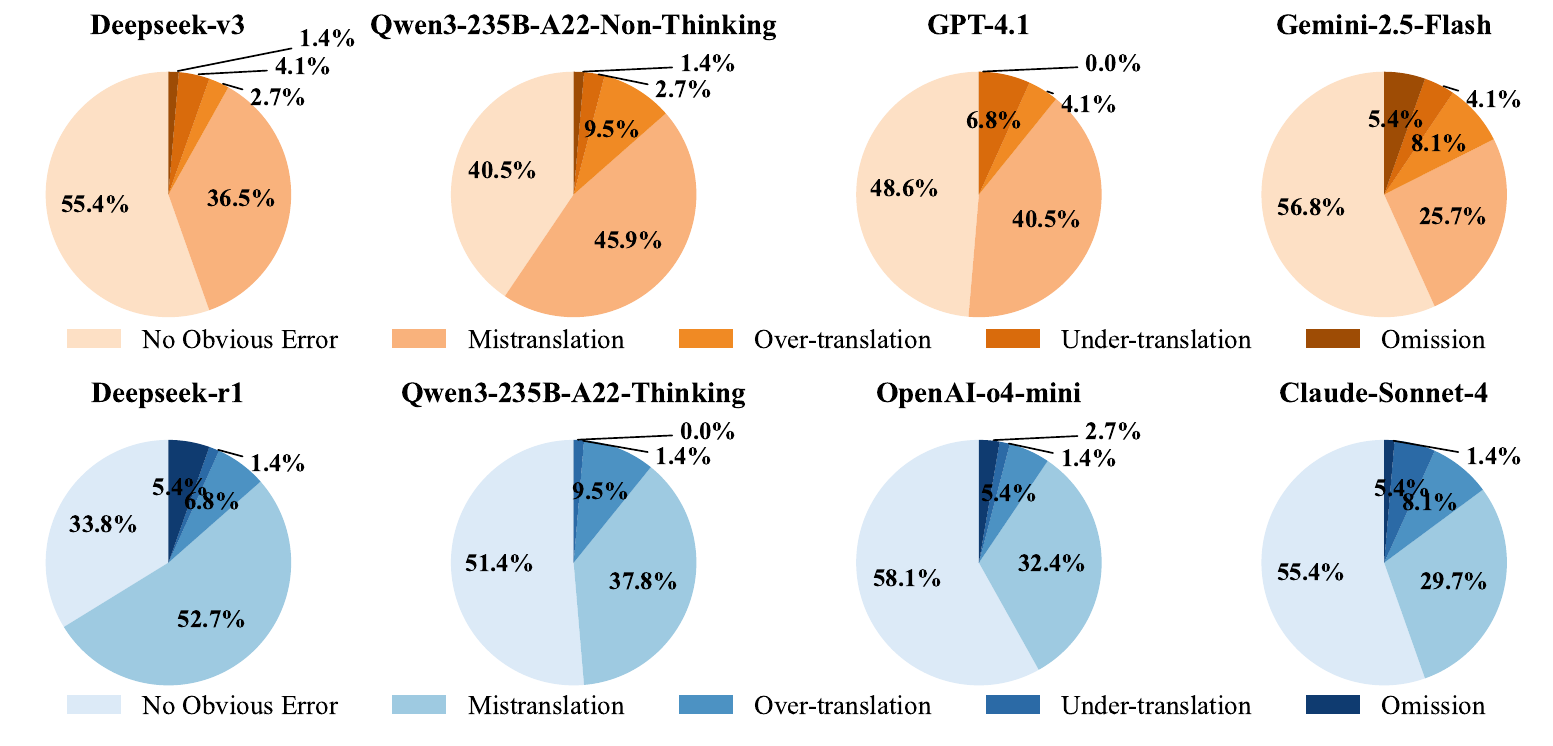}
\vspace{-0.15in}
\caption{Distribution of translation error types across eight models.}
\label{fig:cont-acc}
\end{figure*}

\begin{table*}[t]
\caption{Translation strategy types and corresponding descriptions.}
\vspace{-0.1in}
\centering
\footnotesize
\renewcommand\arraystretch{1.1}
\setlength{\tabcolsep}{1.8mm}{
\resizebox{2\columnwidth}{!}{
\begin{tabular}{lp{10cm}}

\toprule
\textbf{Category} & \textbf{Description} \\
\midrule
{\bf A. Domestication} & Oriented toward the norms of the target culture, adapting the translation to fit the cognitive and cultural expectations of target readers. Strategies include localization, generalization, cultural substitution, functional equivalence, etc.
\\
\addlinespace
{\bf B. Foreignization} & Preserves the cultural characteristics and heterogeneity of the source text, allowing the translation to reflect the original cultural context. Strategies include literal translation, transliteration, semantic borrowing, formal equivalence, etc. \\
\addlinespace
{\bf C. Hybrid Strategies} & Combine domestication and foreignization to strike a balance between cultural fidelity and reader comprehension. Strategies include literal translation with annotations, retention with explanation, partial literal translation combined with partial free translation, etc. \\
\addlinespace
{\bf D. Others} & Translation approaches that do not fall directly under domestication or foreignization, often involving omission or entirely creative renditions. Strategies include omission, innovative translation, etc. \\
\bottomrule

\end{tabular}}}
\label{tab:strategy-type}
\vspace{-0.in}
\end{table*}

\begin{figure*}[t]
\centering
\includegraphics[width=2\columnwidth]{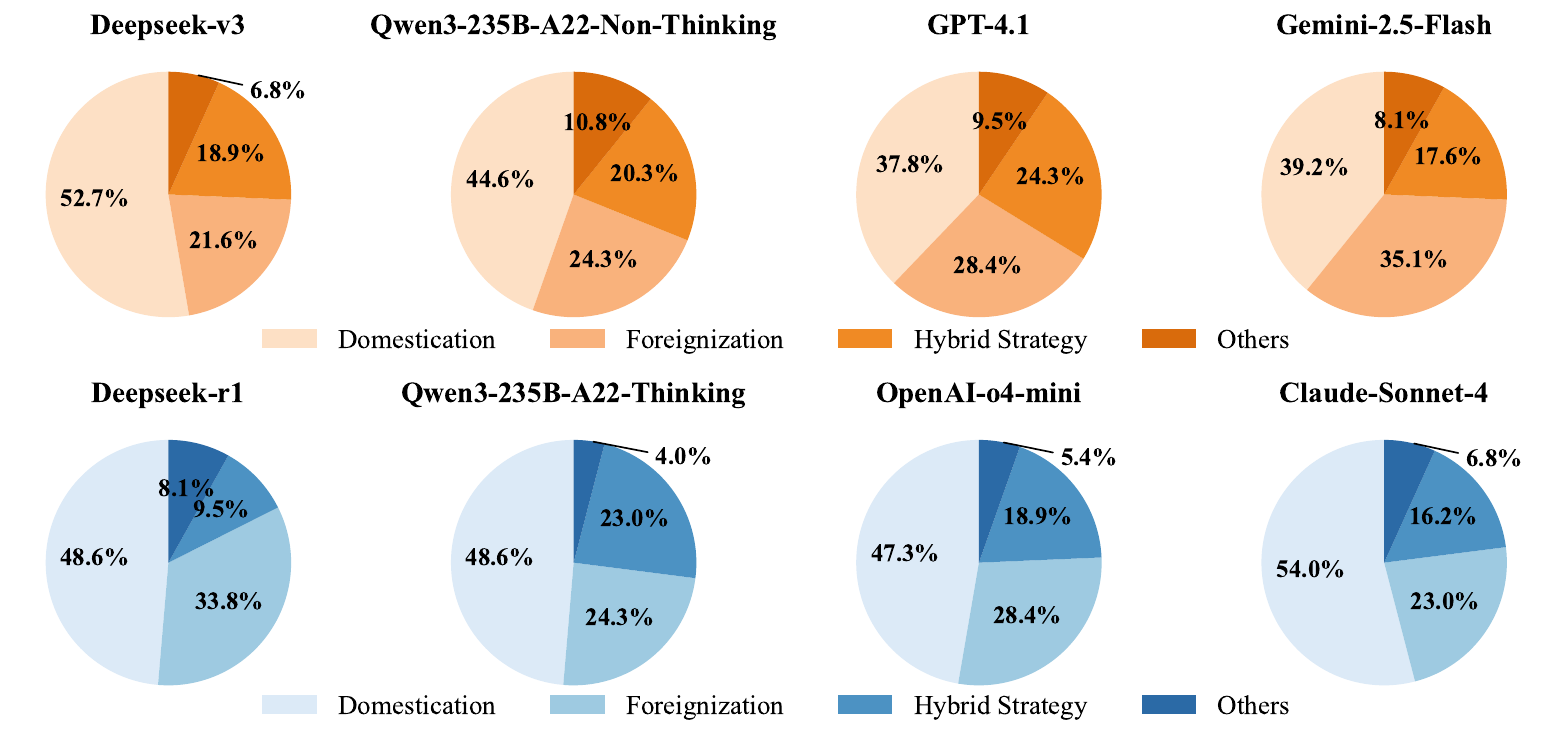}
\vspace{-0.15in}
\caption{Distribution of culture-oriented translation strategy types across eight models.}
\label{fig:trans-strategy}
\end{figure*}

\section{Error Type Analysis}
\label{appe:error-analysis}

To better understand error patterns of models, we analyze error distributions across eight models in \textcolor{deepred}{Tab.\ref{fig:cont-acc}}. Errors are categorized into five types: \textit{No Obvious Error}, \textit{Mistranslation}, \textit{Overtranslation}, \textit{Undertranslation}, and \textit{Omission} (see \textcolor{deepred}{Tab.\ref{tab:error-type}} for details).
Top-performing models, including OpenAI-o4-mini, Gemini-2.5-Flash, and Deepseek-v3, show the highest \textit{No Obvious Error} rates (55–58\%) and lowest \textit{Mistranslation} (25–36\%). In contrast, Deepseek-r1 achieves only 33.8\% error-free and 52.7\% mistranslation, confirming that misunderstanding source content is the primary accuracy bottleneck.

The remaining three error types are relatively rare, together accounting for under 15\% of cases for most models. 
\textit{Undertranslation} and \textit{Omission} remain consistently low across all systems ($<$7\% and $<$5.4\% respectively), indicating models rarely refuse to translate or drop content entirely.
In addition, \textit{Overtranslation} shows an interesting pattern: reasoning models exhibit notably higher rates (Qwen3-235B-A22-Thinking 9.5\%, Claude-Sonnet-4 8.1\%, Gemini-2.5-Flash 8.1\%) compared to non-reasoning Deepseek-v3 (2.7\%). While longer CoT encourages elaboration, such additions often over-explain cultural elements or introduce irrelevant content, failing to improve outcomes. This confirms that ``thinking more'' does not guarantee better cultural adaptation --- effective translation requires appropriately calibrated elaboration.

\section{Translation Strategy Analysis}
\label{appe:strategy-analysis}

We further analyze the translation strategies adopted by the models during the translation process, as shown in \textcolor{deepred}{Fig.\ref{fig:trans-strategy}}.
Specifically, we draw on the ``domestication–foreignization'' theory introduced in \textcolor{deepred}{§\ref{sec:theory}} and categorize translation strategies into four types (see \textcolor{deepred}{Tab.\ref{tab:strategy-type}} for details).
Overall, each model employs either domestication or foreignization strategies in over 80\% of cases, with mixed or other strategies being relatively infrequent.
A clear preference for domestication emerges across models, prioritizing adaptation to target language norms. Notably, reasoning models adopt domestication at higher rates than their non-reasoning counterparts.

Two primary domestication patterns are observed. The first involves Japanese idiomatic expressions, which preserve core meaning and ensure fluency but replace source cultural imagery with target-culture equivalents, resulting in cultural loss. The second relies on generic paraphrasing, which captures only coarse semantics while omitting nuanced components, leading to partial meaning degradation (see \textcolor{deepred}{§}\ref{case:1} and \ref{case:4} for a typical case).
These findings suggest that while domestication enhances readability and target-language acceptability, it often comes at the cost of source culture preservation --- underscoring a key limitation of current models in conveying culturally embedded meanings.

\newpage

\section{Case Study}
\label{appe:case-study}

\subsection{Case 1}
\label{case:1}
\

\noindent\textit{\textbf{Source:}}\\
\begin{CJK}{UTF8}{gbsn}
他又成了香饽饽了，都抢不到手。
\end{CJK}
\vspace{0.2cm}

\noindent\textit{\textbf{Reference:}}\\
\begin{CJK}{UTF8}{min}
あのひとはほかほか饅頭（人気者の意）なってしまい、奪い合いでめったに手にははいらないね
\end{CJK}
\vspace{0.2cm}

\noindent\textit{\textbf{Deepseek-v3:}}\\
\begin{CJK}{UTF8}{min}
彼はまた人気者になって、みんな手に入れようにも手に入れられないんだ。彼はまた人気者になって、みんなが奪い合っている。
\end{CJK}
\vspace{0.2cm}

\noindent\textit{\textbf{Qwen3-235B-A22B-Non-Thinking:}}\\
\begin{CJK}{UTF8}{min}
彼はまたもてはやされる存在になり、手に入れようとしても手に入らないほどだ。
\end{CJK}
\vspace{0.2cm}

\noindent\textit{\textbf{GPT-4.1:}}\\
\begin{CJK}{UTF8}{min}
また引っ張りだこになって、誰も手に入れられないよ。
\end{CJK}
\vspace{0.2cm}

\noindent\textit{\textbf{Gemini-2.5-Flash:}}\\
\begin{CJK}{UTF8}{min}
彼はまた引っ張りだこになって、もう手に入らないよ。
\end{CJK}
\vspace{0.2cm}

\noindent\textit{\textbf{Deepseek-r1:}}\\
\begin{CJK}{UTF8}{min}
彼はまた引く手あまたになってしまって、もらえないんです。
\end{CJK}
\vspace{0.2cm}

\noindent\textit{\textbf{Qwen-235B-A22B-Thinking:}}\\
\begin{CJK}{UTF8}{min}
彼はまたもてはやされる存在となり、誰もが手に入れようとするものの、なかなか手に入らない状態です。
\end{CJK}
\vspace{0.2cm}

\noindent\textit{\textbf{OpenAI-o4-mini:}}\\
\begin{CJK}{UTF8}{min}
彼はまた人気者になって、みんな手に入れようにも手に入れられないんだ。
\end{CJK}
\vspace{0.2cm}

\noindent\textit{\textbf{Claude-sonnet-4:}}\\
\begin{CJK}{UTF8}{min}
彼はまた人気者になって、取り合いになっているよ。
\end{CJK}
\vspace{0.2cm}

\paragraph{Analysis.}
The expression \begin{CJK}{UTF8}{gbsn}“香饽饽”\end{CJK} in the original text functions as a key culturally loaded word. Literally, it refers to a ``steaming hot bun'', while metaphorically it conveys the cultural meaning of ``a highly sought-after person''.
Ito renders it as \begin{CJK}{UTF8}{min}「ほかほか饅頭（人気者の意）」\end{CJK}. This approach represents a classical paradigm in cultural translation: the source-language imagery is preserved through literal translation, while the metaphorical meaning is made explicit through explanatory notes in parentheses.
In this way, the translation both maintains the cultural imagery of the source text and ensures comprehensibility for target-language readers.

LLMs tend to prioritize alignment with the linguistic conventions and expression patterns of the target language at the expense of source-language cultural imagery. This preference is realized primarily through two approaches:

\begin{itemize}[leftmargin=0.4cm]
    \item The first approach involves replacing the source expression with established Japanese idioms. For example, GPT-4.1 and Gemini-2.5-Flash translate the term as \begin{CJK}{UTF8}{min}「引っ張りだこ」\end{CJK}, while Deepseek-r1 uses \begin{CJK}{UTF8}{min}「引く手あまた」\end{CJK}. Although these expressions accurately convey the core meaning of ``being highly sought after'' and produce fluent, idiomatic output, they essentially constitute cultural substitution. As a result, the culturally specific imagery associated with \begin{CJK}{UTF8}{gbsn}“饽饽”\end{CJK}, a symbol rooted in Chinese culinary culture, is lost.
    \item The second approach generalizes the expression into more neutral vocabulary. Examples include \begin{CJK}{UTF8}{min}「人気者」\end{CJK} in the translations by Deepseek-v3 and Claude-4-sonnet, and \begin{CJK}{UTF8}{min}「もてはやされる存在」\end{CJK} produced by Qwen series. While such renderings convey the attribute of “being popular,” they substantially weaken or even eliminate the dynamic semantic component of ``being competed for'' (i.e., the sense of rivalry or competition implied in the original expression). Consequently, part of the original semantic richness is lost.
\end{itemize}

\subsection{Case 2}
\label{case:2}
\

\noindent\textit{\textbf{Source:}}\\
\begin{CJK}{UTF8}{gbsn}
那宝玉是个丈八的灯台，照见人家，照不见自家的。
\end{CJK}
\vspace{0.2cm}

\noindent\textit{\textbf{Reference:}}\\
\begin{CJK}{UTF8}{min}
どだい肝腎の宝玉さまが、『丈八（一丈八尺）のお灯明、他人は照らせても、わが身は照らせぬ（「灯台もと暗し」）』。
\end{CJK}
\vspace{0.2cm}

\noindent\textit{\textbf{Deepseek-v3:}}\\
\begin{CJK}{UTF8}{min}
宝玉は八尺の灯台で、他人を照らすが、自分の足元は照らさない。
\end{CJK}
\vspace{0.2cm}

\noindent\textit{\textbf{Qwen3-235B-A22B-Non-Thinking:}}\\
\begin{CJK}{UTF8}{min}
その宝玉というやつは、丈八（じょうはち）の灯台（とうだい）のようなもので、他人（ひと）のことは照らし出すが、自分自身は照らし出せないのだ。
\end{CJK}
\vspace{0.2cm}

\noindent\textit{\textbf{GPT-4.1:}}\\
\begin{CJK}{UTF8}{min}
あの宝玉は、まるで丈八（じょうはち）の灯台のようなもので、人のことはよく見えるが、自分のことは見えないのです。
\end{CJK}
\vspace{0.2cm}

\noindent\textit{\textbf{Gemini-2.5-Flash:}}\\
\begin{CJK}{UTF8}{min}
この宝玉（ほうぎょく）は、まさに「灯台下暗し（とうだいもとくらし）」という男だった。
\end{CJK}
\vspace{0.2cm}

\noindent\textit{\textbf{Deepseek-r1:}}\\
\begin{CJK}{UTF8}{min}
宝玉は一丈八尺の灯籠台、他人を照らすが、自分の足元は見えぬ
\end{CJK}
\vspace{0.2cm}

\noindent\textit{\textbf{Qwen-235B-A22B-Thinking:}}\\
\begin{CJK}{UTF8}{min}
宝玉は丈八（じょうはち）の灯台のようなものだ。他人は照らすが、自分自身は照らせぬ。
\end{CJK}
\vspace{0.2cm}

\noindent\textit{\textbf{OpenAI-o4-mini:}}\\
\begin{CJK}{UTF8}{min}
あの宝玉は丈八の燈台みたいなもので、人のことはよく照らせても、自分の家（身内）のことはまったく見えない。
\end{CJK}
\vspace{0.2cm}

\noindent\textit{\textbf{Claude-sonnet-4:}}\\
\begin{CJK}{UTF8}{min}
その宝玉は、他人のことはよく分かるが、自分のことは分からない人なのである。
\end{CJK}
\vspace{0.2cm}

\paragraph{Analysis.}
This case belongs to a citation-type two-part allegorical saying \begin{CJK}{UTF8}{gbsn}(歇后语)\end{CJK} structure composed of ``tenor + vehicle + explanation''.
The first half establishes the metaphor, while the second half reveals the intended meaning of the pun. From a translation perspective, the major difficulty of this sentence lies in two aspects: the treatment of the cultural image and the explicit rendering of the metaphorical implication. On the one hand, the translation needs to reproduce the exaggerated metaphorical image of \begin{CJK}{UTF8}{gbsn}“丈八的灯台”\end{CJK}; on the other hand, it must convey the satirical meaning embedded in the original expression, namely ``seeing others’ faults while failing to examine one’s own''.

Ito’s translation adopts a strategy of preserving the source metaphor while providing explanatory support. First, the expression \begin{CJK}{UTF8}{min}「丈八（一丈八尺）のお灯明」\end{CJK} retains the original imagery, and the term \begin{CJK}{UTF8}{gbsn}“丈八”\end{CJK} is annotated to ensure comprehension for target-language readers. Second, the latter half is translated literally as \begin{CJK}{UTF8}{min}「他人は照らせても、わが身は照らせぬ」\end{CJK}, which accurately conveys the meaning of the original while maintaining a formal correspondence with the source structure. Finally, the translator supplements the Japanese proverb \begin{CJK}{UTF8}{min}「灯台もと暗し」\end{CJK}, further facilitating reader understanding. As a result, the translation both preserves the cultural imagery and makes the metaphorical meaning explicit, achieving a desirable balance between cultural fidelity and reader comprehensibility.

For LLMs, their translations can generally be categorized into the following types:

\begin{itemize}[leftmargin=0.4cm]
    \item The first type combines structural preservation with explanatory adaptation. For example, GPT-4.1 adopts a canonical Japanese comparative construction for expressing metaphor and contrast, \begin{CJK}{UTF8}{min}「まるで…のようなもので、…よく見えるが、…見えないのです」\end{CJK}. The resulting translation is fluent and rhythmic, effectively restoring the syntactic structure of the original sentence.
    \item The second type adopts a literal translation strategy for the cultural image. Examples include deepseek-v3-250324, o4-mini-2025-04-16, deepseek-r1-250528, and the qwen3 series. A common characteristic of these translations is that \begin{CJK}{UTF8}{gbsn}“丈八”\end{CJK} is directly rendered as \begin{CJK}{UTF8}{min}「八尺」\end{CJK}, \begin{CJK}{UTF8}{min}「丈八」\end{CJK}, or \begin{CJK}{UTF8}{min}「一丈八尺」\end{CJK}. Although such literal translation preserves the cultural features of the source language to the greatest extent, the absence of further explanation or adaptation may reduce readability for Japanese readers unfamiliar with the cultural reference.
    \item The third type adopts a domestication or paraphrasing strategy, as seen in the translations produced by gemini-2.5-flash and claude-sonnet-4-20250514. While these translations achieve high fluency, the cultural imagery of the source text is entirely replaced or erased. From the perspective of cultural transmission, this strategy may fulfill communicative purposes but inevitably results in a loss of cultural information.
    \item Finally, some translations generated by the DeepSeek series contain clear errors. In these cases, \begin{CJK}{UTF8}{gbsn}“丈八的灯台”\end{CJK} is interpreted as a literal physical object rather than a metaphor, and the metaphorical relationship is therefore not reflected. Moreover, \begin{CJK}{UTF8}{gbsn}“照不见自家”\end{CJK} is rendered as \begin{CJK}{UTF8}{min}「自分の足元は見えない」\end{CJK}, which concretizes the expression into a spatial description. Consequently, the original satirical implication—referring to a lack of self-reflection—degenerates into a neutral depiction of physical visibility.
\end{itemize}

\subsection{Case 3}
\label{case:3}
\

\noindent\textit{\textbf{Source:}}\\
\begin{CJK}{UTF8}{gbsn}
......只听里面称三赞四，耍笑之音虽多；又兼着恨五骂六，忿怨之声亦不少。
\end{CJK}
\vspace{0.2cm}

\noindent\textit{\textbf{Reference:}}\\
\begin{CJK}{UTF8}{min}
......なかでは、ほめたりたたえたりで、笑いさんざめく声がしきりにする一方、またわめいたり恨んだりの怒りと憎しみの声も少なくないふう......
\end{CJK}
\vspace{0.2cm}

\noindent\textit{\textbf{Deepseek-v3:}}\\
\begin{CJK}{UTF8}{min}
中では三人を褒め四人を誉め、笑い戯れる声が多いかと思えば、また五人を恨み六人を罵る、怒り怨む声も少なくない。
\end{CJK}
\vspace{0.2cm}

\noindent\textit{\textbf{Qwen3-235B-A22B-Non-Thinking:}}\\
\begin{CJK}{UTF8}{min}
中からは三人称えて四人ほめる声、遊び笑う音は多いにあり、また五人を恨み六人を罵る声、怒りと不満の言葉も少なくない。
\end{CJK}
\vspace{0.2cm}

\noindent\textit{\textbf{GPT-4.1:}}\\
\begin{CJK}{UTF8}{min}
中からは、「三を褒め四を称える」声や、ふざけ合う音が多く聞こえてきたが、一方では「五を恨み六を罵る」怒りや恨みの声も少なからず混じっていた。
\end{CJK}
\vspace{0.2cm}

\noindent\textit{\textbf{Gemini-2.5-Flash:}}\\
\begin{CJK}{UTF8}{min}
(30)中からはお世辞を言ったり、褒めそやしたり、ふざけ合ったりする声が多く聞こえたが、その一方で、罵声や恨み言、不平不満の声も少なくなかった。
\end{CJK}
\vspace{0.2cm}

\noindent\textit{\textbf{Deepseek-r1:}}\\
\begin{CJK}{UTF8}{min}
中からは賑やかな笑い声とともに、褒めそやす話し声が多く聞こえる一方で、恨み罵る憤りの声も少なからず聞こえてきた。
\end{CJK}
\vspace{0.2cm}

\noindent\textit{\textbf{Qwen-235B-A22B-Thinking:}}\\
\begin{CJK}{UTF8}{min}
中をうかがうと、そこには三人称賛し、四人で笑い声を立てている者が多く、また五人を恨み、六人を罵るような怒りや不満の声も少なくなかった。
\end{CJK}
\vspace{0.2cm}

\noindent\textit{\textbf{OpenAI-o4-mini:}}\\
\begin{CJK}{UTF8}{min}
中からは絶え間なくおだてるような笑い声が聞こえてきたが、同時に罵倒や恨み言を口にする怒りの声も少なくなかった。
\end{CJK}
\vspace{0.2cm}

\noindent\textit{\textbf{Claude-sonnet-4:}}\\
\begin{CJK}{UTF8}{min}
中からは三を褒め四を讃えるような、戯れ笑う声が多く聞こえてくるものの、同時に五を恨み六を罵るような、憤懣怨嗟の声も少なくなかった。
\end{CJK}
\vspace{0.2cm}

\paragraph{Analysis.}
Taking the expression \begin{CJK}{UTF8}{gbsn}“称三赞四”“恨五骂六”\end{CJK}, a typical culturally loaded word under the linguistic category, as an example, the key challenge in its translation lies not only in correctly interpreting the rhetorical device of numerical vagueness characteristic of Chinese, but also in reproducing the rhythmic vitality and emotional intensity created by the parallel structure of the original expression.

Ito renders \begin{CJK}{UTF8}{gbsn}“称三赞四”\end{CJK} as \begin{CJK}{UTF8}{min}「ほめたりたたえたり」\end{CJK} and \begin{CJK}{UTF8}{gbsn}“恨五骂六”\end{CJK} as \begin{CJK}{UTF8}{min}「わめいたり恨んだりの怒り」\end{CJK}. This translation employs the Japanese parallel construction \begin{CJK}{UTF8}{min}「〜たり〜たり」\end{CJK}, which not only accurately conveys the semantic content of the original but also reproduces its rhythmic pattern and emotional tension.
In contrast, LLMs generally fall into two categories.

\begin{itemize}[leftmargin=0.4cm]
    \item On the one hand, the translations by Deepseek-v3, GPT-4.1, and Claude-sonnet-4 mistakenly interpret the numerals \begin{CJK}{UTF8}{gbsn}“三、四、五、六”\end{CJK}(``three, four, five, six''), which function rhetorically as vague intensifiers, as literal quantity markers. Consequently, they are rendered as expressions such as \begin{CJK}{UTF8}{min}「三人を褒め四人を讃える」\end{CJK} (“praising three people and commending four”) or \begin{CJK}{UTF8}{min}「五人を恨み六人を罵る」\end{CJK} (``hating five people and insulting six''), resulting in a clear distortion of the original meaning.
    \item On the other hand, OpenAI-o4-mini, Deepseek-r1, and Gemini-2.5-Flash, render \begin{CJK}{UTF8}{gbsn}“称三赞四”\end{CJK} and \begin{CJK}{UTF8}{gbsn}“恨五骂六”\end{CJK} as expressions such as \begin{CJK}{UTF8}{min}「お世辞を言う」\end{CJK}, \begin{CJK}{UTF8}{min}「褒めそやす」\end{CJK}, \begin{CJK}{UTF8}{min}「罵倒する」\end{CJK}, or \begin{CJK}{UTF8}{min}「恨み言」\end{CJK}. Although these translations successfully convey the core semantic content of praise and denunciation expressed in the original, they lack the linguistic tension and rhetorical dynamism of the source text, and therefore fail to fully reproduce its tone and emotional force.
\end{itemize}

\subsection{Case 4}
\label{case:4}
\

\noindent\textit{\textbf{Source:}}\\
\begin{CJK}{UTF8}{gbsn}
巧媳妇做不出没米的粥来
\end{CJK}
\vspace{0.2cm}

\noindent\textit{\textbf{Reference:}}\\
\begin{CJK}{UTF8}{min}
遣繰り上手の嫁さんでも米なしでは粥はできぬ
\end{CJK}
\vspace{0.2cm}

\noindent\textit{\textbf{Deepseek-v3:}}\\
\begin{CJK}{UTF8}{min}
巧婦、米なしに粥を作ることはできない。
\end{CJK}
\vspace{0.2cm}

\noindent\textit{\textbf{Qwen3-235B-A22B-Non-Thinking:}}\\
\begin{CJK}{UTF8}{min}
巧妇も米がなければ粥は作れない。
\end{CJK}
\vspace{0.2cm}

\noindent\textit{\textbf{GPT-4.1:}}\\
\begin{CJK}{UTF8}{min}
どんなに腕のいい主婦でも、米がなければお粥は作れない。
\end{CJK}
\vspace{0.2cm}

\noindent\textit{\textbf{Gemini-2.5-Flash:}}\\
\begin{CJK}{UTF8}{min}
元手がなければ商売はできない
\end{CJK}
\vspace{0.2cm}

\noindent\textit{\textbf{Deepseek-r1:}}\\
\begin{CJK}{UTF8}{min}
ない袖は振れぬ
\end{CJK}
\vspace{0.2cm}

\noindent\textit{\textbf{Qwen-235B-A22B-Thinking:}}\\
\begin{CJK}{UTF8}{min}
巧婦も米がなければ粥は炊けぬ。
\end{CJK}
\vspace{0.2cm}

\noindent\textit{\textbf{OpenAI-o4-mini:}}\\
\begin{CJK}{UTF8}{min}
どんなに器用な嫁でも、米のないお粥は作れない。
\end{CJK}
\vspace{0.2cm}

\noindent\textit{\textbf{Claude-sonnet-4:}}\\
\begin{CJK}{UTF8}{min}
米なしでは、どんなに腕の良い嫁でも粥は作れない
\end{CJK}
\vspace{0.2cm}

\paragraph{Analysis.}
Overall, the translations produced by LLMs can be broadly divided into two categories.

\begin{itemize}[leftmargin=0.4cm]
    \item The first category adopts a domestication strategy aimed at achieving functional equivalence. A representative example is the DeepSeek-r1 translation \begin{CJK}{UTF8}{min}「ない袖は振れぬ」\end{CJK}. This type of translation completely abandons the concrete imagery in the original text—such as \begin{CJK}{UTF8}{gbsn}“米”\end{CJK}, \begin{CJK}{UTF8}{gbsn}“粥”\end{CJK}, and \begin{CJK}{UTF8}{gbsn}“媳妇”\end{CJK}—and instead aligns directly with a semantically equivalent idiomatic expression in Japanese culture. Although this approach diverges significantly from the original in terms of linguistic form, it succeeds in achieving pragmatic equivalence at the functional level.
    \item The second category adopts a foreignization strategy, preserving the cultural imagery of the source language. For instance, the Qwen3-thinking translation \begin{CJK}{UTF8}{min}「巧婦も米がなければ粥は炊けぬ」\end{CJK} retains the original imagery of \begin{CJK}{UTF8}{gbsn}“巧妇”\end{CJK}, \begin{CJK}{UTF8}{gbsn}“米”\end{CJK}, and \begin{CJK}{UTF8}{gbsn}“粥”\end{CJK}. Moreover, by employing expressions such as \begin{CJK}{UTF8}{min}「炊けぬ」\end{CJK}, which carry a strong classical Sino-Japanese stylistic flavor, the translation closely mirrors the structural and rhetorical style of the original Chinese proverb, thereby achieving a dual form of foreignization in both imagery and syntax.
\end{itemize}

It is worth noting, however, that when foreignization lacks sufficient cultural contextual support, it may lead to semantic deviation. Models such as Gemini-2.5-Flash and GPT-4.1 exhibit varying degrees of strategic fluctuation in their sentence-level translations. For example, Gemini renders the proverb as \begin{CJK}{UTF8}{min}「元手がなければ商売はできぬ」\end{CJK} (“without capital, business cannot be conducted”). Although the translation maintains the formal structure of a proverb, it introduces a substantial shift in meaning by abstracting the original expression into a commercial context.

\subsection{Case 5}
\label{case:5}
\

\noindent\textit{\textbf{Source:}}\\
\begin{CJK}{UTF8}{gbsn}
那畸角上我有一个花冢。
\end{CJK}
\vspace{0.2cm}

\noindent\textit{\textbf{Reference:}}\\
\begin{CJK}{UTF8}{min}
あの隅の方に、わたくし、花塚をこしらえてあります
\end{CJK}
\vspace{0.2cm}

\noindent\textit{\textbf{Deepseek-v3:}}\\
\begin{CJK}{UTF8}{min}
あの隅っこに私の花塚があります。
\end{CJK}
\vspace{0.2cm}

\noindent\textit{\textbf{Qwen3-235B-A22B-Non-Thinking:}}\\
\begin{CJK}{UTF8}{min}
その変な角のところに、私はお墓参りをしています。
\end{CJK}
\vspace{0.2cm}

\noindent\textit{\textbf{GPT-4.1:}}\\
\begin{CJK}{UTF8}{min}
その歪んだ角に、私は一つの花塚（はなづか）を持っている。
\end{CJK}
\vspace{0.2cm}

\noindent\textit{\textbf{Gemini-2.5-Flash:}}\\
\begin{CJK}{UTF8}{min}
その奇妙な角の上に、私は一つの花塚を建てた。
\end{CJK}
\vspace{0.2cm}

\noindent\textit{\textbf{Deepseek-r1:}}\\
\begin{CJK}{UTF8}{min}
あの隅っこに私の花の塚があるんだ。
\end{CJK}
\vspace{0.2cm}

\noindent\textit{\textbf{Qwen-235B-A22B-Thinking:}}\\
\begin{CJK}{UTF8}{min}
あの片隅に、私の花塚があります。
\end{CJK}
\vspace{0.2cm}

\noindent\textit{\textbf{OpenAI-o4-mini:}}\\
\begin{CJK}{UTF8}{min}
あの歪んだ角に、私はひとつの花の塚を持っている。
\end{CJK}
\vspace{0.2cm}

\noindent\textit{\textbf{Claude-sonnet-4:}}\\
\begin{CJK}{UTF8}{min}
その隅っこに私は花塚を持っています。
\end{CJK}
\vspace{0.2cm}

\paragraph{Analysis.}
\begin{CJK}{UTF8}{gbsn}“畸角”\end{CJK} originally refers to a spatial notion, meaning a ``remote, neglected, or unnoticed corner''. The translation strategies adopted by LLMs models can be broadly divided into two categories.

\begin{itemize}[leftmargin=0.4cm]
    \item The first category consists of contextually grounded literary translations. Representative examples include the Reference translation, Qwen3-235B-A22B-Thinking, and the Deepseek series. These models correctly identify the spatial attribute embedded in the character \begin{CJK}{UTF8}{gbsn}“畸”\end{CJK} within Chinese rhetoric. The Reference translation uses \begin{CJK}{UTF8}{min}「隅の方」\end{CJK}, while DeepSeek renders it as \begin{CJK}{UTF8}{min}「隅っこ」\end{CJK}, both of which accurately anchor the expression in physical spatial orientation. Among them, the translation \begin{CJK}{UTF8}{min}「片隅」\end{CJK} produced by Qwen3-235B-A22B-Thinking stands out as particularly effective. In Japanese lexical intuition, \begin{CJK}{UTF8}{min}「片隅」\end{CJK} not only denotes a marginal space but also carries a subtle sense of loneliness and isolation, which aesthetically resonates with the imagery of \begin{CJK}{UTF8}{gbsn}“花冢”\end{CJK}. This alignment suggests that models equipped with stronger reasoning abilities can move beyond literal lexical correspondence and achieve cross-lingual equivalence at the level of scene or atmosphere, rather than merely word meaning.
    \item The second category can be described as moderate translations based on literal semantic equivalence. Models such as GPT-4.1, OpenAI-o4-mini, and Gemini-2.5-Flash exhibit a clear tendency toward dictionary-like literal translation. They interpret the character \begin{CJK}{UTF8}{gbsn}“畸”\end{CJK} as indicating abnormal shape, rendering the phrase as expressions such as \begin{CJK}{UTF8}{min}「歪んだ角」\end{CJK} or \begin{CJK}{UTF8}{min}「奇妙な角」\end{CJK}. Similarly, Qwen3-non-thinking translates it as \begin{CJK}{UTF8}{min}「変な角」\end{CJK}. Such treatments establish only a superficial lexical correspondence and fail to capture the intended meaning of the original expression.
\end{itemize}

In summary, the quality of translating \begin{CJK}{UTF8}{gbsn}“畸角”\end{CJK} largely depends on whether the model recognizes its metaphorical function as an environmental modifier rather than a description of shape. Models capable of detecting contextual cues and employing expressions such as \begin{CJK}{UTF8}{min}「片隅」\end{CJK} or \begin{CJK}{UTF8}{min}「隅」\end{CJK}, which align with natural usage in the target language, demonstrate clear advantages over literalist approaches in both translational accuracy and literary aesthetic quality.

\end{document}